\documentclass[lettersize,journal]{IEEEtran}
\usepackage{subfigure}
\usepackage{amsmath,amsfonts}
\usepackage{algorithmic}
\usepackage{algorithm}
\usepackage{array}
\usepackage[caption=false,font=normalsize,labelfont=sf,textfont=sf]{subfig}
\usepackage{textcomp}
\usepackage{stfloats}
\usepackage{url}
\usepackage{verbatim}
\usepackage{graphicx}
\usepackage{cite}
\usepackage{amsthm}
\usepackage{bm}
\DeclareMathOperator*{\argmin}{argmin}
\newtheorem{theorem}{Theorem}
\newtheorem{corollary}{Corollary}[theorem]

\newtheorem{remark}{Remark}
\usepackage{booktabs}
\usepackage{physics}
\begin{document}

\title{Exploiting Latent Properties to Optimize Neural Codecs}

\author{Muhammet Balcilar, Bharath Bhushan Damodaran,  Karam Naser, Franck Galpin, Pierre Hellier
\thanks{The authors are at InterDigital Inc., France. (email: {muhammet.balcilar, bharath.damodaran}@interdigital.com, pierre.hellier@inria.fr)\\ © 2024 IEEE. Personal use of this material is permitted. Permission
from IEEE must be obtained for all other uses, in any current or future media,
including reprinting/republishing this material for advertising or promotional
purposes, creating new collective works, for resale or redistribution to servers
or lists, or reuse of any copyrighted component of this work in other works.}}

\markboth{IEEE Transactions on Image Processing}%
{Balcilar \MakeLowercase{\textit{et al.}}: Exploiting Latent Properties to Optimize Neural Codecs}

\IEEEpubid{This article has been accepted for publication in IEEE Transactions on Image Processing. DOI 10.1109/TIP.2024.3522813, © 2024 IEEE}

\maketitle

\begin{abstract}
End-to-end image and video codecs are becoming increasingly competitive, compared to traditional compression techniques that have been developed through decades of manual engineering efforts. These trainable codecs have many advantages over traditional techniques, such as their straightforward adaptation to perceptual distortion metrics and high performance in specific fields thanks to their learning ability. However, current state-of-the-art neural codecs do not fully exploit the benefits of vector quantization and the existence of the entropy gradient in decoding devices. In this paper, we propose to leverage these two properties (vector quantization and entropy gradient) to improve the performance of off-the-shelf codecs. Firstly, we demonstrate that using non-uniform scalar quantization cannot improve performance over uniform quantization. We thus suggest using predefined optimal uniform vector quantization to improve performance. Secondly, we show that the entropy gradient, available at the decoder, is correlated with the reconstruction error gradient, which is not available at the decoder. We therefore use the former as a proxy to enhance compression performance. Our experimental results show that these approaches save between $1$ to $3\%$ of the rate for the same quality across various pre-trained methods. In addition, the entropy gradient based solution improves traditional codec performance significantly as well.  
\end{abstract}

\begin{IEEEkeywords}
Compression, Neural codec, Entropy model.
\end{IEEEkeywords}

\section{Introduction}
\label{sec:intro}
\IEEEPARstart{L}{ossy} image and video compression is a fundamental task in image processing, which became crucial in the time of the pandemic and the increasing volume of video streaming.
Thanks to the community's decades long efforts, traditional methods (e.g. Versatile Video Coding (VVC)) have reached current state of the art rate-distortion (RD) performances and dominate the current codecs market. Recently, end-to-end trainable deep models with promising RD performances have emerged thanks to informative latent learning and the latent distribution modeling. Even though deep learning-based models clearly exceed many traditional techniques and surpass human capability for general computer vision tasks, they are only slightly better than the best traditional codecs for single image compression, according to our knowledge. 

End-to-end deep compression methods typically refer to rate-distortion auto-encoders \cite{habibian2019video}, in which the latent representation is generated by jointly optimizing the encoder, decoder, and entropy model with a rate-distortion loss function. For perceptual compression, distortion based on a perceptual metric can also be used in the loss function \cite{blau19a}. These methods can be seen as a special case of Variational Autoencoder (VAE) models as described in \cite{kingma2013auto}, where the approximate posterior distribution is a uniform distribution centered on the encoder's outputs (latents) at the training stage, and has a fixed variance output distribution and trainable priors \cite{theis2017lossy,balle2016end}. It was shown that minimizing the evidence lower bound (ELBO) of this special VAE is equivalent to jointly minimizing the mean square error (MSE) of the reconstruction and the entropy of the latents w.r.t the priors \cite{balle2018variational}. All proposed models mainly differ in the way they model the priors: using either fully-factorized \cite{balle2016end}, zero-mean Gaussian \cite{balle2018variational}, Gaussian \cite{minen_joint,minnen2020channel} or a mixture of Gaussian \cite{cheng2020image} distribution models. Some methods predict the priors using an autoregressive schema \cite{minen_joint,minnen2020channel,cheng2020image,xie2021enhanced, he2021checkerboard}, some improve them through global and local context modeling \cite{qian21, Kim_2022_CVPR} or transformer based architecture \cite{liu2023learned}. These neural image codecs were extended to video compression by using two VAEs, one for encoding motion information and another for encoding residual information in end-to-end video compression \cite{dvc,ssf,aivc,lhbdc,pourreza2021extending, LI_deep_context,lidcvc}. 
\IEEEpubidadjcol

An important step in building a neural codec is the quantization of the latents before entropy coding. Nearly all of the mentioned prior state-of-the-art models use a fixed bin-width uniform Scalar Quantization (SQ). Although Vector Quantization (VQ) is theoretically better \cite{gersho2012vector}, 
there have been very few attempts to use VQ in neural codecs.  
For instance \cite{agustsson2017soft,multi_gauss_VQ2022,feng2023nvtc} learn non-uniform  partitioning map where the convergence of these grids might encounter some difficulties, such as being highly sensitive to initial partitioning, collapsing of some partitions, existence of unused partitions during training and also high search complexity to find nearest partitioning during encoding. Thus, it may require many tricks that are defined in \cite{feng2023nvtc}. Uniform VQ such as implemented in \cite{kudo2023lvqvaeendtoend,zhang2023lvqac} can solve above mentioned problems by using predefined uniform partitioning. However \cite{multi_gauss_VQ2022,feng2023nvtc,kudo2023lvqvaeendtoend} learn complex encoder/decoder network and inter-correlation between VQ dimensions that results a new and better entropy model and \cite{zhang2023lvqac} couples VQ with a new custom transformation. Thus, their performance improvements are not because of purely better quantization and their VQs are not applicable on top of existing neural codecs with SQ.

A general principle in compression is to exploit all available information at the decoder to reconstruct the data. Surprisingly, even though the entropy gradients w.r.t latents are available at the decoder, this information remains unused so far in the literature. Similar works in the literature have attempted to improve the performance of the codec during encoding, for example by using specific parameterization \cite{reducinggap}, or computationally heavy fine-tuning solutions \cite{ NEURIPS2020_066f182b, guo21c}, either partially \cite{Campos_2019_CVPR_Workshops,LuCZCOXG20} or entirely \cite{van2021overfitting}. However, all of these methods disregard the existence of the entropy gradient in decoding.%

In this paper, we present two novel contributions that aim at leveraging the properties of learned latent representations for compression. Firstly, we demonstrate that no kind of non-uniform scalar quantization can improve performance compared to uniform scalar quantization, if the neural model is expressive enough. We thus discarded scalar quantization and propose to use uniform vector quantization over the latents. Since the optimal uniform VQ map is known up to certain dimensions, learning partitioning is not needed. This contribution can be applied even without re-training the model if the original model is trained for uniform SQ, which is often the case for neural codecs. 

Secondly, we applied the Karush–Kuhn–Tucker (KKT) conditions to the neural codec, which had not yet been done, to the best of our knowledge except in our recent publication \cite{10222469}, where we gave formal proofs and extended analysis in this paper. These conditions reveal connection between the reconstruction error gradient (unavailable at the decoder) and the entropy gradient (available at the decoder). This finding motivated us to test correlations between these two gradients. We experimentally found a strong correlation between the two gradients for various neural codecs. We therefore used the available entropy gradient as a proxy for the unavailable reconstruction gradient to improve the performance of neural codecs without requiring re-training. These two contributions are generic enough (as they do not depend on the encoder-decoder architecture) to achieve a rate saving of 1-3\% for the same quality and for several neural codec architectures. Last but not least, we showed that our entropy gradient based solution can improve performance of traditional video codec by $0.1\%$ as well  in  \cite{jvetae0125}. Our solution is adopted to the Enhanced Compression Model (ECM) version 10.0 by Joint Video Experts Team (JVET).

\section{Problem statement and State of the Art}
\label{sec:problem_statement}

\begin{figure}
\begin{center}    \includegraphics[width=.47\textwidth]{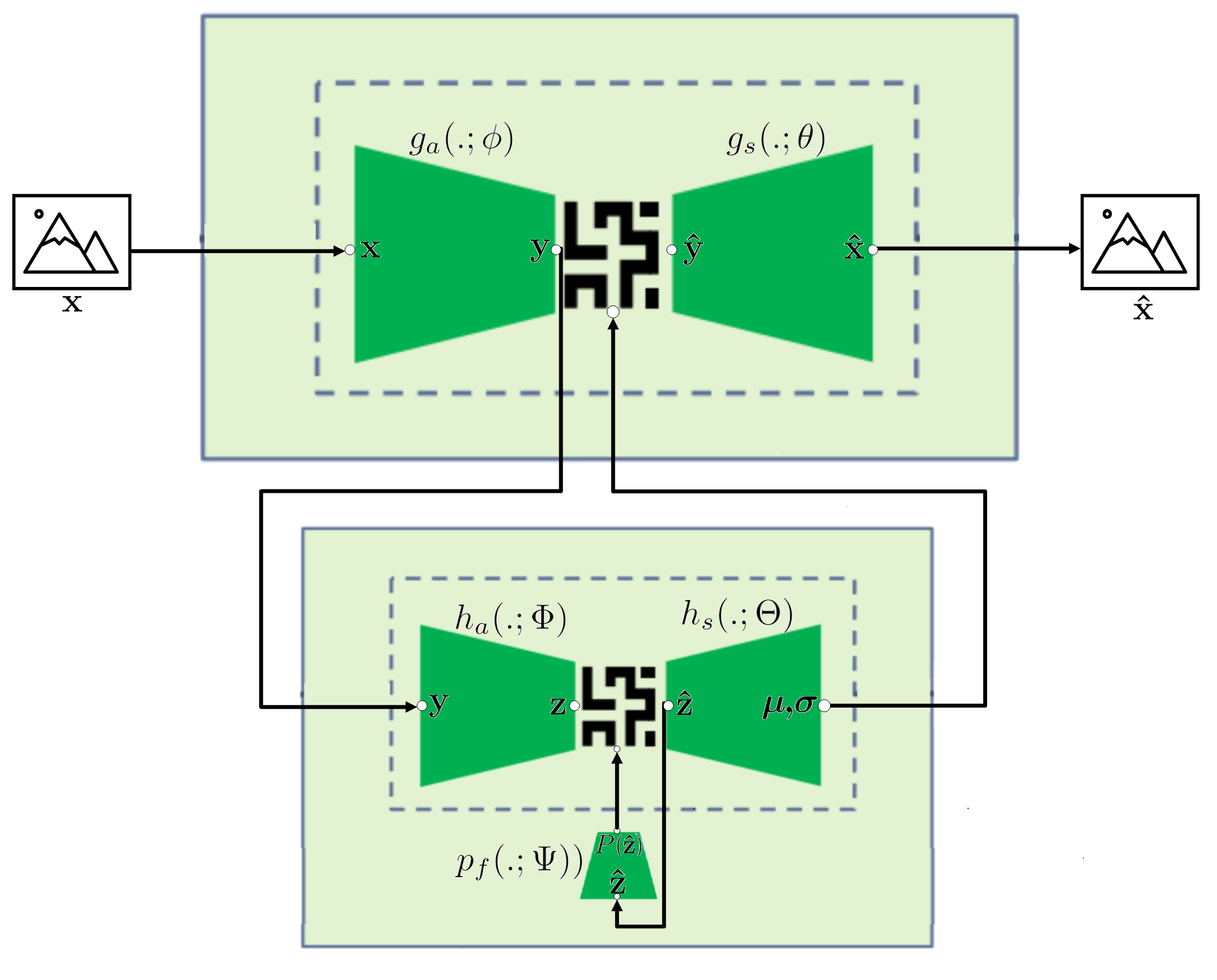}
\end{center}
\caption{Block diagram of the state-of-the art neural codecs. The five dark green blocks are the trainable blocks implemented by neural networks, while the binary patterns show the quantization and entropy encoding/decoding processes driven by certain entropy model's PMFs on main and side latents. 
}
\label{fig:end2end}
\end{figure}

As shown in Figure \ref{fig:end2end}, given an input color image $\mathbf{x} \in \mathbb{R}^{n \times n \times 3}$ to be compressed (the image can be considered to be square without any loss of generality), the neural codec learns a non-linear encoder $g_a (\mathbf{x};\mathbf{\phi})$, parameterized by weights $\mathbf{\phi}$. The encoder output, $\mathbf{y} \in \mathbb{R} ^{m \times m \times o}$, is called the main embeddings (or main latents) of the image. The latent representation is then quantized as $\mathbf{\Tilde{y}} =Q(\mathbf{y})$ to obtain the main codes of the image. De-quantization block $\mathbf{\hat{y}}=Q^{-1}(\mathbf{\Tilde{y}})$ is used to obtain the reconstructed main latents $\mathbf{\hat{y}}$ at the decoding. Decompressed image  $\mathbf{\hat{x}} \in \mathbb{R}^{n \times n \times 3}$ is obtained using the trained deep decoder with $\mathbf{\hat{x}}=g_s(\mathbf{\hat{y}};\mathbf{\theta})$. 
The neural codec is trained to minimize two objectives simultaneously, namely the distortion between $\mathbf{x}$ and $\mathbf{\hat{x}}$, and the length of the bitstream needed to encode $\mathbf{\Tilde{y}}$. The codes are losslessly encoded into a bitstream using an entropy encoder such as Range Asymmetric Numeral Systems (RANS) \cite{duda2009asymmetric}. RANS requires the Probability Mass Function (PMF) of each code, which is also learned at the training stage. Since RANS is asymptotically optimal, the lower bound of bitlength according to Shanon's entropy theorem can be used, instead of the experimental bitlength from RANS, in order to make the bitlength objective differentiable. Thus, neural codecs use the entropy model to learn the PMF of each codes under predefined quantization method, which allows us to determine the lower bound of the bitlength.

Current state-of-the-art neural codecs use a hyperprior entropy model, where the side embeddings (or side latents)  $\mathbf{z} \in \mathbb{R} ^{t \times t \times s}$ are learned by another deep neural network with $\mathbf{z}=h_a(\mathbf{y};\Phi)$. 
The side embeddings are quantized as $\mathbf{\Tilde{z}} =Q(\mathbf{z})$ to obtain side codes $\mathbf{\Tilde{z}}$, followed by de-quantization $\mathbf{\hat{z}}=Q^{-1}(\mathbf{\Tilde{z}})$ to obtain reconstructed side latents $\mathbf{\hat{z}}$. The main motivation of side information is to remove any image structure that would persist in the main latent representation $y$. 
The hyperprior entropy model assumes that the probability density function (PDF) of each scalar latent follows a Gaussian distribution, where the parameters are obtained using another deep network such as $\bm{\mu,\sigma}=h_s(\mathbf{\hat{z}};\Theta)$) \footnote{In autoregressive prediction, they are predicted step by step using the previously decoded main latents as $\bm{\mu_i,\sigma_i}=h_s(\mathbf{\hat{z},\mathbf{\hat{y}}_{<i}};\Theta)$}. Thus, the prediction of the hyperprior model can be defined as $p_{\mathbf{\hat{y}}_{ijk}}(.):=\mathcal{N}(.|\bm{\mu}_{ijk},\bm{\sigma}_{ijk})$. The PMF of latent code $P(\mathbf{\Tilde{y}}_{ijk})$ can be written under predefined quantization method as a function of $p_{\mathbf{\hat{y}}_{ijk}}(.)$. Using this PMF, the lower bound of the main codes' bitlength can be defined as $-\log(p_h(\mathbf{\hat{y}};\mathbf{\hat{z}},\Theta)):=-\sum_{ijk} \log(P(\mathbf{\Tilde{y}}_{ijk}))$. The factorized entropy model learns the PDF for each $t \times t$ latent slice defined as $p_{\mathbf{\hat{z}}_{:,:,s}}(.)$. This PDF is sufficient to compute the PMF of side codes $P(\mathbf{\hat{z}}_{ijk})$ under predefined quantization method. The lower bound of side codes' bitlength can thus be defined by $-\log(p_f(\mathbf{\hat{z}};\Psi)):=-\sum_{ijk} \log(P(\mathbf{\Tilde{z}}_{ijk}))$.

In this setting, the deep encoder ($g_a (.;\mathbf{\phi})$), deep decoder ($g_s(.;\mathbf{\theta})$), hyperprior entropy model 
$p_h(.;\mathbf{\hat{z}},\Theta)$ (composed of deep hyperprior encoder ($h_a(.;\Phi)$), deep hyperprior decoder ($h_s(.;\Theta)$) and factorized entropy model $p_f(.;\Psi)$) are the trainable blocks implemented by neural networks. Each block with its trainable parameter, input and output are depicted in  Figure \ref{fig:end2end}.
The optimal values of parameters $\mathbf{\phi},\mathbf{\theta},\Phi,\Theta$ and $\Psi$ are found by minimizing the following loss function using training samples

\begin{equation}
   \label{eq:loss}
   \mathcal{L}=\mathop{\mathbb{E}}_{\substack{\mathbf{x}\sim p_x }}\left[-\log(p_f(\mathbf{\hat{z}};\Psi))-\log(p_h(\mathbf{\hat{y}};\mathbf{\hat{z}},\Theta))  + \lambda d(\mathbf{x},\mathbf{\hat{x}})\right],
\end{equation}
where $d(.,.)$ is any distortion measure between the original and the reconstructed image (for example the mean squared error). The rate term (r) is the sum of the lower bound of the bitlength of the side information ($-\log(p_f(\mathbf{\hat{z}};\Psi))$) and the main information ($-\log(p_h(\mathbf{\hat{y}};\mathbf{\hat{z}},\Theta))$). Hyper-parameter $\lambda$ controls the trade-off between the rate (r) and distortion (d) terms.

Both the quantization step $Q(.)$ and its de-quantization $Q^{-1} (.)$ counterpart need to be applied to the main and side information. In addition, both the factorized and hyperprior entropy models need to know about the predefined quantization technique to obtain the PMF. To the best of our knowledge, most of the current methods implement quantization as a 1-bin width uniform Scalar Quantization (SQ) method, with a few exceptions: \cite{agustsson2017soft,agustsson2020universally,guo21c}. 
This  quantization step is implemented by element-wise nearest integer rounding $Q(x)=round(x)$ and its de-quantization counterpart $Q^{-1}(x)=x$. The PMF of $\mathbf{\tilde{y}}_{ijk}\in \mathbb{R}$ can thus be calculated by  $P(\mathbf{\tilde{y}}_{ijk})=\int_{\mathbf{\hat{y}}_{ijk}-0.5}^{\mathbf{\hat{y}}_{ijk}+0.5}p_{\mathbf{\hat{y}}_{ijk}}(x)dx$ where $p_{\mathbf{\hat{y}}_{ijk}}(.)$ is the PDF of latent $\mathbf{\hat{y}}_{ijk}$ learnt by the entropy model \footnote{The entropy model can alternatively learn cumulative distribution function CDF, $\sigma_{\mathbf{\hat{y}}_{ijk}}(x)$ instead of PDF and calculate PMF by $P(\mathbf{\tilde{y}}_{ijk})=\sigma_{\mathbf{\hat{y}}_{ijk}}(\mathbf{\hat{y}}_{ijk}+0.5)-\sigma_{\mathbf{\hat{y}}_{ijk}}(\mathbf{\hat{y}}_{ijk}-0.5)$.}. Since the nearest integer rounding operation has non-informative gradients, a continuous relaxation must be applied at the training stage as $Q(x)=x+\epsilon$, where $\epsilon$ is randomly sampled from uniform distribution $\epsilon \sim U(-0.5,0.5)$. 

However, in the case of Vector Quantization (VQ), latents can be packed into a $v$-dimensional vector $u \in \mathbb{R}^{v}$ and each $u$ is assigned to a single code. Quantization centers $\mathbf{c}^{(i)} \in \mathbb{R}^{v}, i=0 \dots M-1$ are learned by the entropy model. Thus, the quantization step finds the index of the nearest center as $Q(x)=\argmin_{i}{||x-\mathbf{c}^{(i)}||}$, while de-quantization returns the quanta center as $Q^{-1}(i)=\mathbf{c}^{(i)}$. In this case, there are $M$ different quantization centers, and $M$ unique codes. Since the $\operatorname{argmin}$ operator applies hard assignment, it has non-informative gradients, and continuous relaxation must also be applied during training. This is generally achieved by a $\operatorname{softmax}$ operator that assigns all codes to the latent vector with different probabilities, depending on the distances to the centers. These probabilities are used by the entropy model to learn the PMF of each quanta center (codes), and also for de-quantization which is the expectation of quanta centers under these probabilities during training.

\section{Uniform Vector Quantization}
\label{sec:vq}
Uniform SQ is widely used in neural codecs, though it is not the optimal quantizer among all SQ methods. In fact, it is known that the optimal quantizer should have smaller grid sizes in regions of higher probability and larger grid sizes in regions of lower probability \cite{farvardin1984optimum}. Thus, a non-uniform SQ method that is aware of source distributions should have a lower quantization error than uniform SQ. However, we state in the following theorem that uniform SQ is sufficient among all SQ methods in the neural codecs. 

\begin{theorem}
  \label{Th:th1}
If a neural codec has an encoder block $g_a: \mathbb{R}^{n \times n \times 3} \rightarrow \mathbb{R}^{m \times m \times o}$, an decoder block $g_s: \mathbb{R}^{m \times m \times o} \rightarrow \mathbb{R}^{n \times n \times 3}$ and it requires a non-uniform SQ map for optimal rate-distortion performance, there exists another neural codec that produces the same rate-distortion performance with 1-bin width uniform SQ (nearest integer rounding quantization) whose encoder block is $f \circ g_a $ and decoder block is $g_s \circ f^{-1}$ where $f: \mathbb{R}^{m \times m \times o} \rightarrow \mathbb{R}^{m \times m \times o}$ is an invertible transformation.
\end{theorem}

\begin{figure*}[t]
  \centering
  \subfigure[] {
    \includegraphics[width=.27\textwidth]{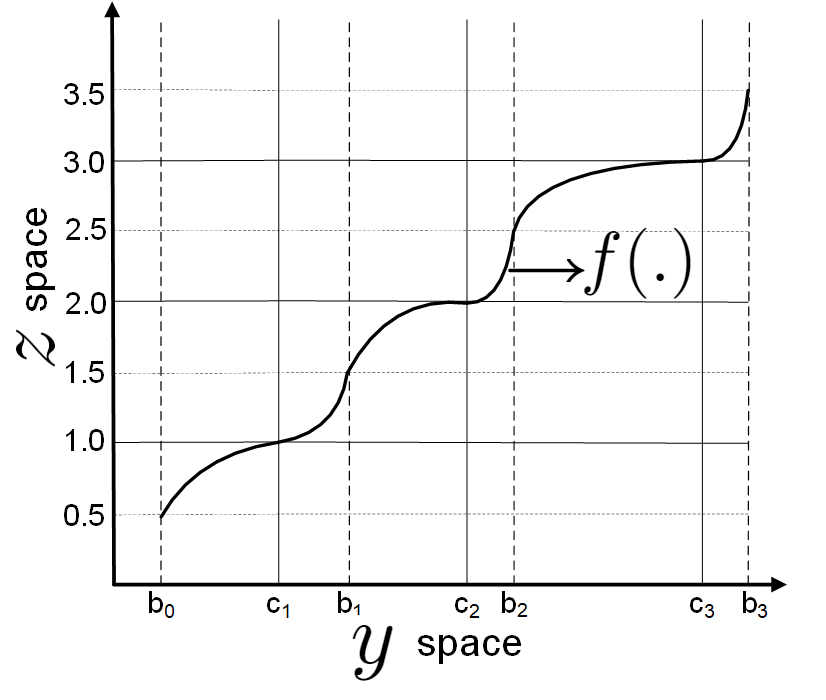}
  }
  \subfigure[] {
    \includegraphics[width=.27\textwidth]{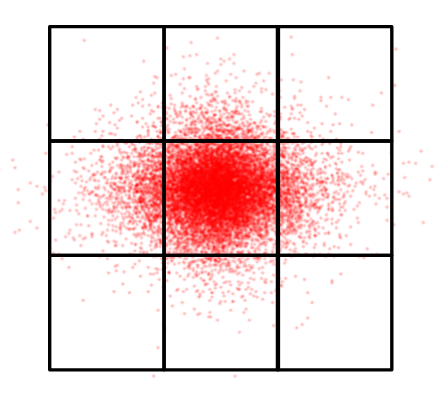}
  }
  \subfigure[] {
    \includegraphics[width=.27\textwidth]{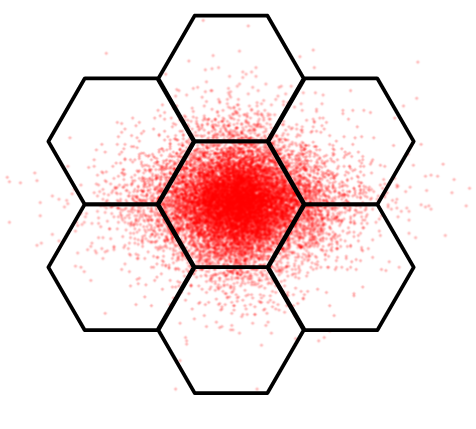}
  }
  \caption{{a) $f: y \rightarrow z $ transforms a non-uniform quantization map (grid borders $b_0,\dots b_n$ and grid centers $c_1,\dots c_n$) into a uniform map (centers are located on integer where borders are at the middle of two consecutive centers)}. b) Uniform SQ grids on 2D  c) Optimal uniform VQ grids on 2D.
  }
\label{fig:VQ1}
\end{figure*}

The proof can be found in Appendix~\ref{section:pth1} and is based on modeling the one-dimensional quantizer using a memoryless monotonically increasing nonlinearity followed by a uniform fixed-point quantizer as in \cite{bennett1948spectra}. We show that an invertible function that can be implemented by neural networks can transform the borders and grid centers of a non-uniform quantization map into a uniform quantization map. A simple illustration of  this transformation
is shown in Figure~\ref{fig:VQ1}(a). Since the transformation is element-wise and smooth, we assume that a shallow neural network with fewer number of parameters can learn this transformation. According to Theorem~\ref{Th:th1}, a neural codecs with nearest integer rounding quantization is sufficient among all scalar quantization methods as long as they are expressive enough to learn this transformation. 
We can assume that expressivity is sufficient, given that both the encoder and decoder are composed of deep neural networks with millions of parameters in the latest neural codecs such as \cite{minen_joint,balle2018variational,cheng2020image,lidcvc,xie2021enhanced,liu2023learned}. The advantage of using invertible layers between less powerful encoder and decoder blocks was recently shown experimentally in \cite{shukor2022video}. In addition, the sufficiency of uniform quantization in neural codecs was discussed in \cite{BalleCMSJAHT21}. Our theorem verifies these two prior contributions.

Since Theorem \ref{Th:th1} excludes non-uniform SQs in neural codecs, a possible remaining selection for lower quantization error can be VQ. Even though VQ is theoretically better than SQ (despite the dimensions are being i.i.d \cite{1056457}), so far it has not been able to perform significantly better than uniform SQ in neural codecs without changing encoder/decoder and entropy model. In the next section we show how VQ can improve off-the-shelf neural codecs performance by replacing SQ.

\subsection{Space Tessellation Grids }

In this paper, we suggest using uniform vector quantization based on space tessellation. This method involves using a predefined high dimensional grid to cover the entire high-dimensional space. The quantization grids are thus fixed instead of learned. The nearest integer quantization grid in 1D is equivalent to a unit square grid in 2D, as shown in Figure~\ref{fig:VQ1}(b) for a zero-mean Gaussian source. However it may not be the best partition of the space with uniform grids, therefore using  nearest integer quantization (i.e square grids in 2D) directly in the neural codec might not yield better RD performances. The following remark shows the existence of entropy constrained  optimal space tessellation and can be used for uniform vector quantization.
\begin{remark}
 \label{rm:rm1}
VQ constraints with a uniform grid produces optimal space tessellation, which refers to a pattern of $v$-dimensional shapes that fit perfectly together without any gaps, and have minimum inertia. The optimal shapes are regular hexagons in the case of 2D, and truncated octahedron for 3D. 
\cite{gersho1979asymptotically}.
\end{remark}
\begin{figure*}[t]
  \centering
  \subfigure[] {
    \includegraphics[width=.3\textwidth]{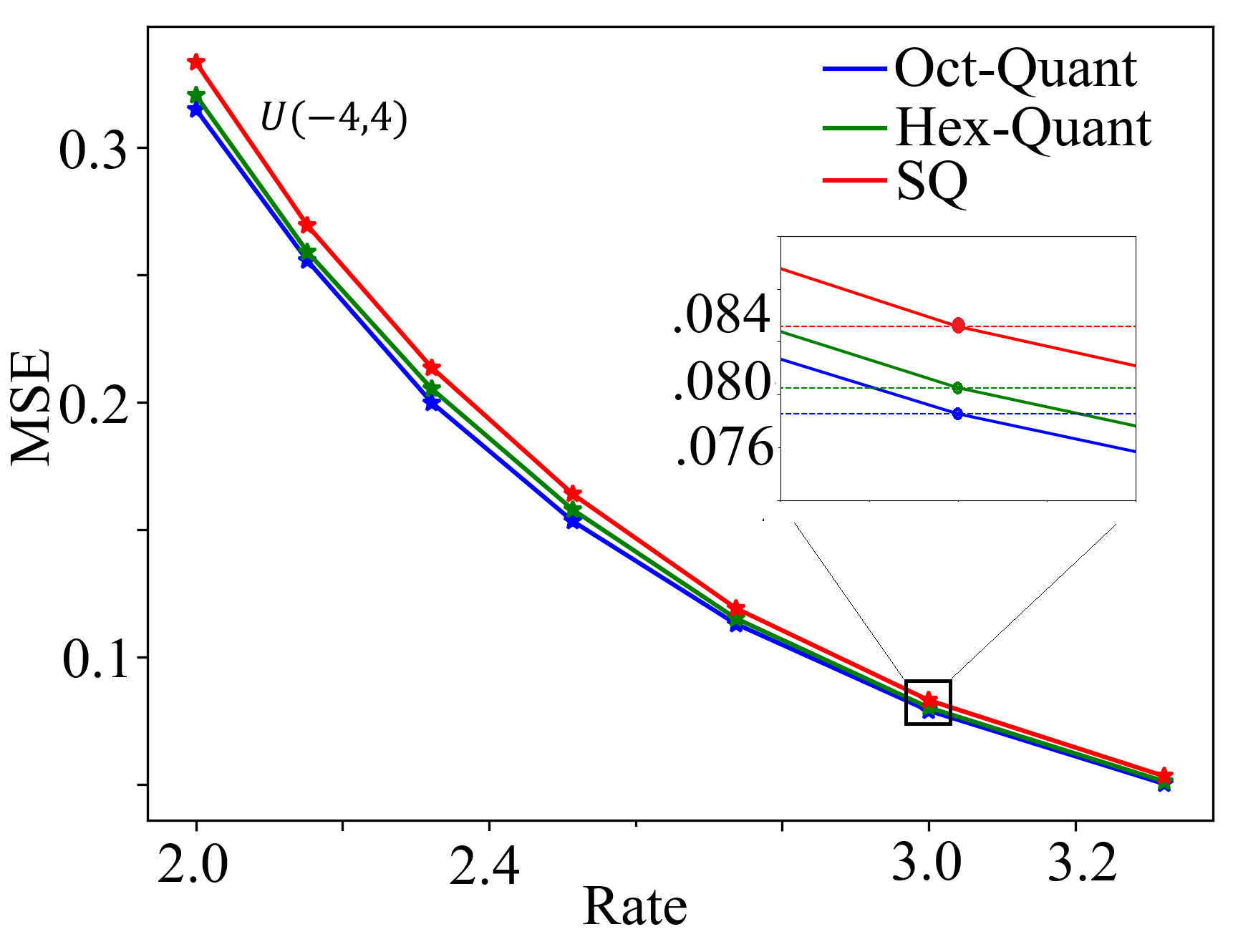}
  }
  \subfigure[] {
    \includegraphics[width=.3\textwidth]{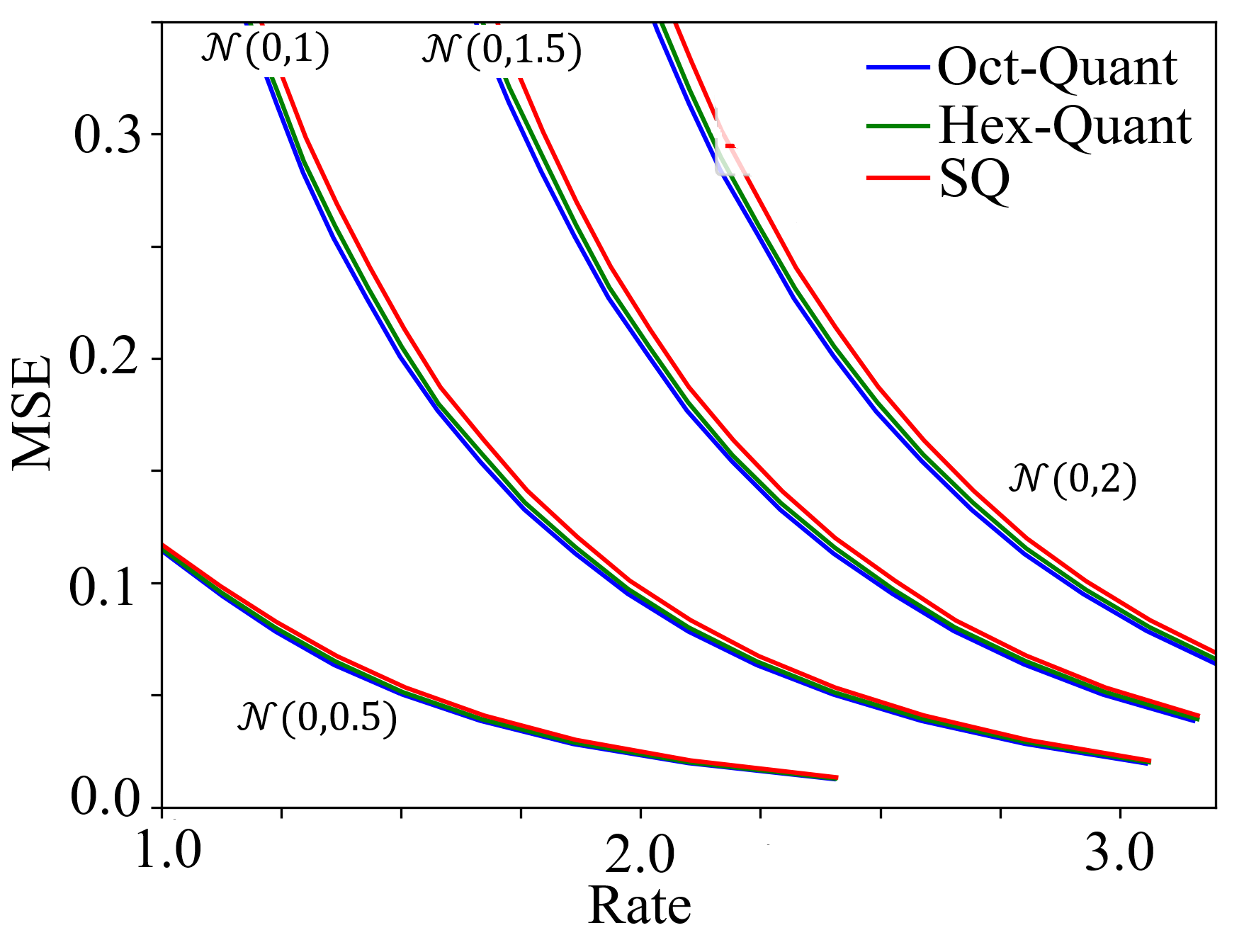}
  }
  \subfigure[] {
    \includegraphics[width=.3\textwidth]{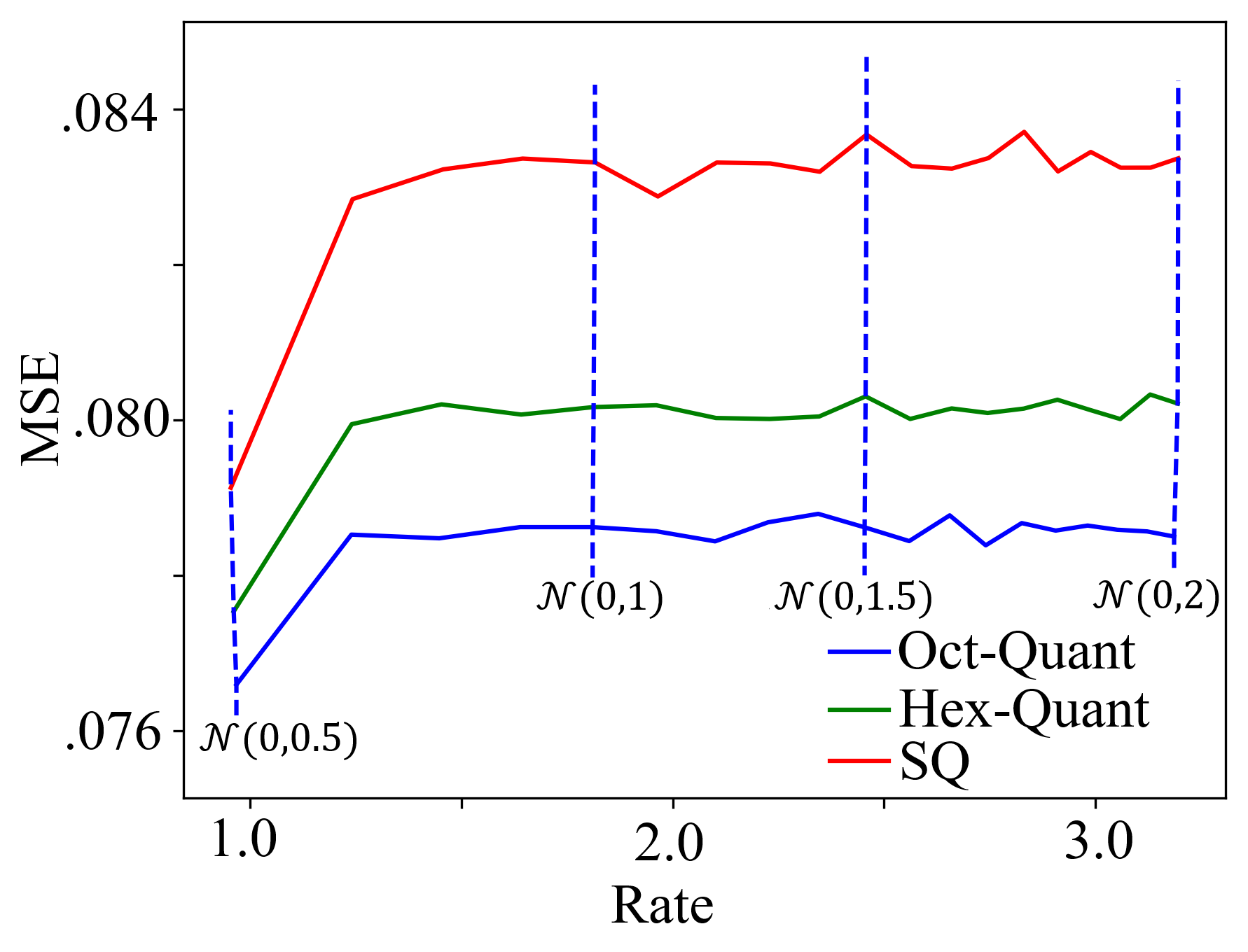}
  }
  \caption{RD performances of different volume uniform SQ, regular hexagon \textbf{Hex-Quant} and truncated octahedron \textbf{Oct-Quant} grid. a) Uniform source is sampled from $U(-4,4)$ and zoom-in where the grid volume is unitary. b) Different Gaussian sources. c) RD plot of unitary volume grids for Gaussian sources.}
\label{fig:VQ_sim}
\end{figure*} 
The above remark indicates that instead of learning the grid centers and optimal partition, we can use the optimal space tessellation grids for the corresponding dimensions. In this paper our proposal for 2D space is to use a regular hexagonal grid with a unitary volume, as shown in Figure ~\ref{fig:VQ1}(c). The theoretical advantage of space tessellation grids compared to nearest integer quantization is shown in Appendix ~\ref{section:hexQuant} for a uniform distribution. 

\textbf{Illustration:} We also showcase the simulation results under uniform and Gaussian sources in Figure~\ref{fig:VQ_sim}a-b. For these simulations, we generated random numbers under a uniform distribution $U(-4,4)$ for Figure~\ref{fig:VQ_sim}a, and four different zero-mean Gaussian distributions with specified scales for Figure~\ref{fig:VQ_sim}b. We quantized these numbers using three different methods: uniform SQ (\textbf{SQ}), regular hexagonal grid (\textbf{Hex-Quant}), and truncated octahedral grid (\textbf{Oct-Quant}), with varying grid sizes. The left-hand side of the $x$-axis represents higher rates where the grid size is small, while the right-hand side represents smaller rates where the grid size is large. In Figure~\ref{fig:VQ_sim}a, we zoom in on the rate where the grid size is unitary. 
The MSE of the three different quantization grids are very close to their theoretical values, which are $0.0833$, $0.0801$, and $0.0785$ respectively, as shown in Appendix \ref{section:hexQuant}. For Figure~\ref{fig:VQ_sim}c, we generated random numbers from a zero-mean Gaussian distribution at different scales in the range of $[0.5,2]$, with a $0.1$ interval between scales. We quantized each source using a unitary volume grid of the three quantization maps mentioned above. The left-hand side of the x-axis represents a lower rate when a source from smaller scale Gaussian distributions is used, while the right-hand side represents a higher rate when a source generated by larger-scale Gaussian distributions is used. For all three methods, a single source does not necessarily have the same rate under a unitary grid size. Thus, the vertical dashed lines representing the four different distributions are not well aligned. This simulation is particularly relevant, because the main information in modern neural codecs is a zero-mean Gaussian distribution at different scale, quantized using unitary grids.  Our simulation demonstrates that, at the same bitrate, the uniform vector quantization (VQ) grid has lower mean squared error (MSE) than SQ. This simulation emphasizes the importance of using uniform VQ as a more effective method for quantization in neural codecs. In the following, we show how our uniform vector quantization can be applied to any off-the-shelf neural codec without requiring re-training. 

\subsection{Uniform VQ with Off-the-shelf Neural Codec}

Let $ \mathbf{y} \in \mathbb{R}^{d}$ be the latents of the deep encoder, $v$ the dimension at which the optimal space tessellation is carried out,
$\mathbf{c}^{(i)} \in \mathbb{R}^{v}, i=0 \dots M-1$ the $M$ grid centers of the $v$-dimensional shape. 
In order to quantize the latents, we reshaped the latents $\mathbf{y}$ into pseudo $v$-dimensional latents such that $ \mathbf{y} \in \mathbb{R}^{d} \rightarrow \mathbf{\check{y}} \in \mathbb{R}^{\frac{d}{v} \times v}$.  
Quantization is performed by assigning a value to the nearest neighbors grid centers, as $\mathbf{\Tilde{y}}_j=Q(\mathbf{\check{y}}_j)=\argmin_{i}{||\mathbf{\check{y}}_j-\mathbf{c}^{(i)}||}$, where $\mathbf{\Tilde{y}}_j \in \{0 \dots M-1\}, j=1 \dots \frac{d}{v}$ are the codes to be encoded into a bitstream. In practice, instead of reshaping all latents together into $v$ dimension, we reshape the latents whose learned PDFs are the same (i.e come from the same distribution). 
This approach does not have significantly different results, but leads to a smaller memory footprint at the decoder side since fewer PMF tables are stored. 

To encode the  codes ($\mathbf{\Tilde{y}}$) into a bit-stream, we cannot use the off-the-shelf neural codec's 1D entropy model's PMF calculation, as our latents are $v$-dimensional latents with different domain. We therefore created the PMF by integrating the pseudo-$v$ dimensional PDF
($\prod_{k=1}^{v}p(u_k)$) into grid $G$ whose center is located on $\mathbf{c}^{(i)}$: 
\begin{equation}
\label{eq:hexpmf}   P(\mathbf{c}^{(i)})=\int_{G+\mathbf{c}^{(i)}}\left(\prod_{k=1}^{v}p(u_k)\right) du. 
\end{equation}

Here $P(\mathbf{c}^{(i)})$ is the PMF of $i$-th symbol in the dictionary and $p:\mathbb{R} \rightarrow \mathbb{R}$ is the PDF of the corresponding latent, learned by the entropy model from the baseline model.
Since there is no closed-form solution for the integral of a multi-dimensional Gaussian distribution over a hexagonal and truncated octahedral domain \cite{savaux2020computation}, we use numerical solvers to approximate the solution of \eqref{eq:hexpmf}. We can compute a numerical solution for the main codes PMFs, where $p(.)$ 
is Gaussian distribution as well as non-parametric $p(.)$ 
for the side codes' PMFs. In Appendix~\ref{section:alg1}, we show how to solve equation~\ref{eq:hexpmf} for the hexagonal domain. Another way to calculate this integral is by using a Monte-Carlo simulation for known PDFs $p(.)$ 
before deployment, which we use for the truncated octahedral domain. 

The de-quantization step is to assign the center of the quantized grid to the reconstructed latent by $\mathbf{\bar{y}}_j=\mathbf{c}^{(\mathbf{\Tilde{y}}_j)}$. At the last step, we need to revert the dimension to the original as $\mathbf{\bar{y}} \in \mathbb{R}^{\frac{d}{v} \times v} \rightarrow  \mathbf{\hat{y}} \in \mathbb{R}^{d}$ which should followed by decoder block to obtain reconstructed image in the general framework.
\section{Forgotten Information: The Entropy Gradient}
\label{sec:shiftlatent}

The receiver can compute the gradients of the main/side entropy w.r.t the main/side latents after decoding the main/side latents. Since these gradients are only available after decoding the latent codes, they may not seem useful at first sight. This could be the reason why these gradients have not been used at inference time so far. Here, we propose to use these gradients through the analysis of the Karush-Kuhn-Tucker conditions, and we experimentally found that they can be correlated with other useful gradients.

\begin{figure}
\begin{center}
    \includegraphics[width=.40\textwidth]{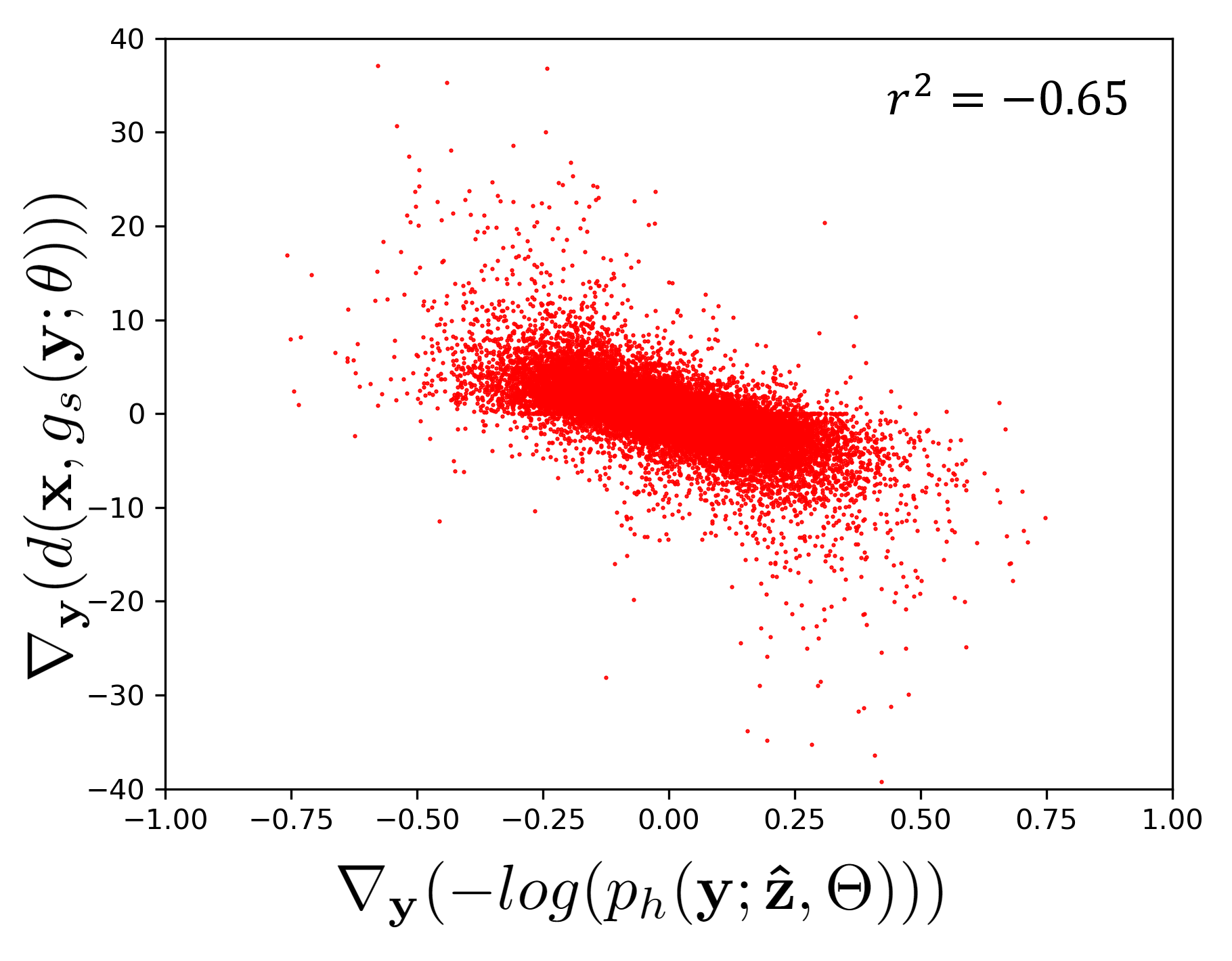}
\end{center}
\caption{Correlation between the gradients of the entropy and of the reconstruction error w.r.t the main latents.
}
\label{fig:corr}
\end{figure}

Let us view the neural codec's loss function in \eqref{eq:loss} as an unconstrained multi-objective optimization problem. The optimal solution of such a problem is called Pareto Optimal: no objective can be made better without making at least one other objective worse  \cite{miettinen2012nonlinear}. The following remark shows a useful property of a solution of unconstrained multi-objective optimization.
\begin{remark}
 \label{rm:kkt}
\textbf{Karush–Kuhn–Tucker (KKT) conditions:}  
If the problem is $w^*=\argmin_{w}(\sum_i \alpha_i.\mathcal{L}_i(w))$;
where $\alpha_i \geq 0$, $\sum \alpha_i = 1$ and $\mathcal{L}_i$ is the $i$-th objective to be minimized, solution $w^*$ is Pareto optimal if and only if it satisfies $\sum_i \alpha_i\nabla_w\mathcal{L}_i(w^*))=0$ \cite{desideri2012multiple}.
\end{remark}
This remark is also valid for end-to-end compression models. The following theorem shows how to leverage the KKT conditions for end-to-end image compression models.
\begin{theorem}
  \label{Th:th2}
If an end-to-end compression model is Pareto Optimal, it satisfies the following conditions.
\begin{equation}
\label{eq:kkt1}
   \mathbb{E}_{\substack{\mathbf{x}\sim p_x }}\left[ \nabla_{\Phi}\log(p_f(\mathbf{\hat{z}};\Psi)) + \nabla_{\Phi}\log(p_h(\mathbf{\hat{y}};\mathbf{\hat{z}},\Theta))     \right]=0.
\end{equation}
\begin{multline}
\label{eq:kkt2}
   \mathbb{E}_{\substack{\mathbf{x}\sim p_x }} \nabla_{\phi}\left[-\log(p_f(\mathbf{\hat{z}};\Psi))-\log(p_h(\mathbf{\hat{y}};\mathbf{\hat{z}},\Theta)) \right] + \\
   \lambda  \nabla_{\phi}d(\mathbf{x},g_s(\mathbf{\hat{y}};\mathbf{\theta})) =0.
\end{multline}
\end{theorem}

Proof is in Appendix~\ref{section:pth2}. The following corollary shows its usefulness under a certain assumption.  

\begin{corollary}
  \label{Th:col1}
If we assume that \eqref{eq:kkt1}  and \ref{eq:kkt2} are valid for every single input $\mathbf{x}$ independently, then an optimal end-to-end compression model satisfies the following conditions:
\begin{equation}
\label{eq:col1}
\nabla_{\mathbf{\hat{z}}}\log(p_f(\mathbf{\hat{z}};\Psi))  =  -\nabla_{\mathbf{\hat{z}}}\log(p_h(\mathbf{\hat{y}};\mathbf{\hat{z}},\Theta)).     
\end{equation}
\begin{multline}
\label{eq:col2}
    \nabla_{\mathbf{\hat{y}}}\left [-\log(p_f(\mathbf{\hat{z}};\Psi))-\log(p_h(\mathbf{\hat{y}};\mathbf{\hat{z}},\Theta)) \right] = \\
    -\lambda  \nabla_{\mathbf{\hat{y}}}d(\mathbf{x},g_s(\mathbf{\hat{y}};\mathbf{\theta})).      
\end{multline}
\end{corollary}

Proof can be found in Appendix~\ref{section:pcol2}. Under this assumption, it implies that the gradient of the main and side information's entropy w.r.t the side latent have opposite directions in \eqref{eq:col1}. Additionally, the weighted reconstruction error gradient and the the total entropy gradient w.r.t the main latent also have opposite directions in \eqref{eq:col2}. Hence, under this assumption, we claim that these pairs of gradients have a correlation coefficient of $-1$ and one of them can be used as a proxy for the other by changing the sign.

The presence of a correlation has been validated for different neural codecs using different test sets, as presented in Appendix~\ref{sec:ablation} and section \ref{sec:exp}. The scatter plot in Figure~\ref{fig:corr} illustrates the correlation between the gradient of the main entropy and the gradient of the reconstruction error w.r.t the main latents for a specific image sample where the gradients were calculated on a neural codec described in \cite{minen_joint}. Based on our tests on various widely-used neural codecs, we experimentally observed relatively strong correlations between the gradients in \eqref{eq:col2}\footnote{Since we found an equivalent correlation, we use it as $\nabla_{\mathbf{\hat{y}}}\left [-\log(p_h(\mathbf{\hat{y}};\mathbf{\hat{z}},\Theta)) \right] = -\lambda  \nabla_{\mathbf{\hat{y}}}d(\mathbf{x},g_s(\mathbf{\hat{y}};\mathbf{\theta}))$.} (ranging from -0.1 to -0.5) and weaker correlations (ranging from -0.15 to 0.1) for the gradients in \eqref{eq:col1}. Even though the experimental correlations differ from $-1$, we believe they are significant enough to be used in practice. This discrepancy can be attributed to the fact that the assumption is only valid on average for the training database, and holds less strongly for any particular sample.

\textbf{Latent Shift wrt Gradients.} Based on the definition of gradient-based optimization, $\mathbf{\hat{z}}$ needs to take a step in the negative direction of $\nabla_{\mathbf{\hat{z}}}(-\log(p_h(\mathbf{\hat{y}};\mathbf{\hat{z}},\Theta)))$ in order to decrease the main information bitlength $-\log(p_h(\mathbf{\hat{y}};\mathbf{\hat{z}},\Theta))$. However $\nabla_{\mathbf{\hat{z}}}(-\log(p_h(\mathbf{\hat{y}};\mathbf{\hat{z}},\Theta)))$ is not available before decoding $\mathbf{\hat{y}}$. On the other hand, the correlated gradient $\nabla_{\mathbf{\hat{z}}}(-\log(p_f(\mathbf{\hat{z}};\Psi)))$ is known after decoding $\mathbf{\hat{z}}$. We claim that there exists a step size $\rho_f^*$ that decrease the bitlength of the main information such that:
\begin{multline}
\notag
   -\log(p_h(\mathbf{\hat{y}};\mathbf{\hat{z}},\Theta)) \geq  \\  -\log(p_h(\mathbf{\hat{y}};\mathbf{\hat{z}}+\rho_f^*\nabla_{\mathbf{\hat{z}}}(-\log(p_f(\mathbf{\hat{z}};\Psi)))),\Theta).
\end{multline}
$\rho_f^*$ can be obtained through numerical search, or by optimization:
\begin{equation}
\label{eq:opt_side}
   \rho_f^*=\argmin_{\rho_f}(-\log(p_h(\mathbf{\hat{y}};\mathbf{\hat{z}}+\rho_f\nabla_{\mathbf{\hat{z}}}(-\log(p_f(\mathbf{\hat{z}};\Psi)))),\Theta)).
\end{equation}
For the second condition, $\mathbf{\hat{y}}$ needs to take a step in the negative direction of $\nabla_{\mathbf{\hat{y}}}(d(\mathbf{x},g_s(\mathbf{\hat{y}};\mathbf{\theta})))$ in order to decrease reconstruction error $d(\mathbf{x},g_s(\mathbf{\hat{y}};\mathbf{\theta}))$. $\nabla_{\mathbf{\hat{y}}}(d(\mathbf{x},g_s(\mathbf{\hat{y}};\mathbf{\theta})))$ is never available during decoding, but the correlated gradient $\nabla_{\mathbf{\hat{y}}}(-\log(p_f(\mathbf{\hat{z}};\Psi))-\log(p_h(\mathbf{\hat{y}};\mathbf{\hat{z}},\Theta)))$ is known after decoding $\mathbf{\hat{y}}$ and $\mathbf{\hat{z}}$.  We also claim that there exists a step size $\rho_h^*$ that decrease the reconstruction error:   
\begin{multline}
\notag
   d(\mathbf{x},g_s(\mathbf{\hat{y}};\mathbf{\theta})) \geq  d(\mathbf{x},g_s(\mathbf{\hat{y}}+\rho_h^*\nabla_{\mathbf{\hat{y}}}(-\log(p_f(\mathbf{\hat{z}};\Psi))- \\
   \log(p_h(\mathbf{\hat{y}};\mathbf{\hat{z}},\Theta)));\mathbf{\theta})).  
\end{multline}
$\rho_h^*$ can be found by numerical search or optimization:
\begin{multline}
\label{eq:opt_side2}
   \rho_h^*=\argmin_{\rho_h}(d(\mathbf{x},g_s(\mathbf{\hat{y}}+\rho_h\nabla_{\mathbf{\hat{y}}}(-\log(p_f(\mathbf{\hat{z}};\Psi))- \\  \log(p_h(\mathbf{\hat{y}};\mathbf{\hat{z}},\Theta)));\mathbf{\theta}))).
\end{multline}
Our proposal can be seen as shifting the side latent by $\mathbf{\hat{z}} \leftarrow \mathbf{\hat{z}}+\rho_f^*\nabla_{\mathbf{\hat{z}}}(-\log(p_f(\mathbf{\hat{z}};\Psi)))$ after decoding $\mathbf{\hat{z}}$ and shifting the main latent by $\mathbf{\hat{y}} \leftarrow \mathbf{\hat{y}}+\rho_h^*\nabla_{\mathbf{\hat{y}}}(-\log(p_f(\mathbf{\hat{z}};\Psi))-\log(p_h(\mathbf{\hat{y}};\mathbf{\hat{z}},\Theta)))$\footnote{Since we found similar correlations between the gradient of the main information's entropy and the gradient of the reconstruction error, we applied it as $\mathbf{\hat{y}} \leftarrow \mathbf{\hat{y}}+\rho_h^*\nabla_{\mathbf{\hat{y}}}(-\log(p_h(\mathbf{\hat{y}};\mathbf{\hat{z}},\Theta)))$ in our tests.} after decoding $\mathbf{\hat{y}}$. The optimal step sizes $\rho_f^*, \rho_h^* \in \mathbb{R}$ are tested out of 8 selections each at the encoding stage and consequently added to the bitstream explicitly using $3$ bits for each step size.

\section{Experimental Results}
\label{sec:exp}
In this section, we show the main results, the complexity analysis and some relevant ablation studies.
\subsection{Main Results}
We used the CompressAI library \cite{compressai} to test our contributions on $4$ pre-trained neural image codecs named \textbf{bmshj2018-factorized} in \cite{balle2016end}, \textbf{mbt2018-mean} and \textbf{mbt2018} in \cite{minen_joint},  \textbf{cheng2020-attn} in \cite{cheng2020image}, $3$ additional pre-trained codec named \textbf{invcompress} in \cite{xie2021enhanced}, \textbf{LIC-TCM} in \cite{liu2023learned} and \textbf{DCVC-intra} in \cite{lidcvc} in all intra mode and two neural video codecs named \textbf{SSF} in \cite{ssf} and \textbf{DCVC} in \cite{lidcvc}. For the evaluation, we used the Kodak dataset \cite{eastman_kodak_kodak_nodate}, Clic-2021 Challenge's dataset \cite{CLIC} and UVG dataset \cite{mercat2020uvg}. The codecs were taken off the shelf without retraining. The rate was calculated based on the final length of the compressed data and RGB PSNR was used for distortion.  
To evaluate the performance, we used the bd-rate metric \cite{bjontegaard2001calculation} that measures the bitrate savings achieved for an overlapped range of distortion level \cite{ecm80ctc}. We conducted our evaluation on 6 to 8 pre-trained models provided by compressAI and the authors official implementation, whose bitrates range from 0.1bpp to 1.6bpp for image codecs and 0.03bpp to 0.35bpp for video codecs.
We reported our uniform VQ proposals under the name of \textbf{Hex-Quant} for regular hexagons and \textbf{Oct-Quant} for truncated octahedrons. We called our gradient-based improvement by \textbf{Latent Shift} and the results using both methods \textbf{Join}.   

\begin{table*}[htbp]
\centering
\caption{Average BD-Rate gains of our proposals for different baseline image codecs on 2 image datasets.}\label{tab:image}
\footnotesize
\begin{tabular}{lcccccc}
\toprule
&  \multicolumn{3}{c}{Kodak Test Set} & \multicolumn{3}{c}{Clic-2021 Test Set}\\
\cline{2-7}
Baseline Codec & Hex-Quant.& Latent Shift & Join & Hex-Quant. & Latent Shift & Join \\
\midrule
bmshj2018-factorized & -0.62\% &-0.49\% & -1.08\% & -1.78\% & -0.69\% & -2.26\%   \\
mbt2018-mean  & -0.88\% & -1.27\% & -2.24\% & -0.76\% & -1.21\% & -2.03\%   \\
mbt2018 & -0.55\% & -1.44\% & -1.98\% & -0.66\% & -1.71\% & -2.33\%  \\
cheng2020-attn  & -0.16\% & -0.46\% & -0.73\% & -0.32\% & -0.72\% & -1.15\%  \\
 InvCompress  &  -0.21\% &	-0.55\%	 &-0.82\% &	-0.52\% &-0.63\% &	-1.12\% \\
DCVC-intra mode  &	-0.97\%	& -0.30\%&	-1.28\%&	-1.22\%	&-0.11\% &	-1.44\%\\
LIC-TCM&	-0.92\%	& -0.15\%&	-1.12\%&	-0.97\%	&-0.10\% &	-1.15\%\\
\bottomrule
\end{tabular}
\vspace{-3pt}
\end{table*}

\noindent \textbf{Results on Image Codecs:} In Table~ \ref{tab:image}, we present the results obtained by our methods on different datasets and codecs. Since our two proposals rely on completely different effects, the gain of the \textbf{Join} method is almost the sum of both methods. The reason why the two effects do not add up exactly is the following: after uniform VQ, the reconstructed latents change compare to uniform SQ. The gradients w.r.t. these latents therefore change as well. This variation creates slightly different results compared to \textbf{Latent Shift}. Since uniform VQ reconstructed latent are much closer to the latents before quantization, in almost all cases \textbf{Latent Shift} with uniform VQ offers a better contribution than \textbf{Latent Shift} alone.

According to results in  Table~ \ref{tab:image}, we have seen that if the method uses autoregressive schema such as \textbf{invcompress} and  \textbf{cheng2020-attn}, \textbf{Hex-Quant} has less gain. This is because of the fact that all latent could not be quantized together, but to be done step by step because of the autoregressive prediction. Within the method does not use autoregressive prediction, \textbf{Hex-Quant} has almost $1\%$ gain. \textbf{Latent Shift}'s gain is highly related to the complexity of the decoder network. The more complicated the decoder, the less correlation between gradients, thus less gain. Since the current tendency of neural codec is to have stronger encoder block but shallow decoder block in order to decrease decoding complexity in a level of traditional codecs, this point will not be problematic in the long run.

\begin{figure*}[htbp]
  \centering
  \subfigure[] {
    \includegraphics[width=.3\textwidth]{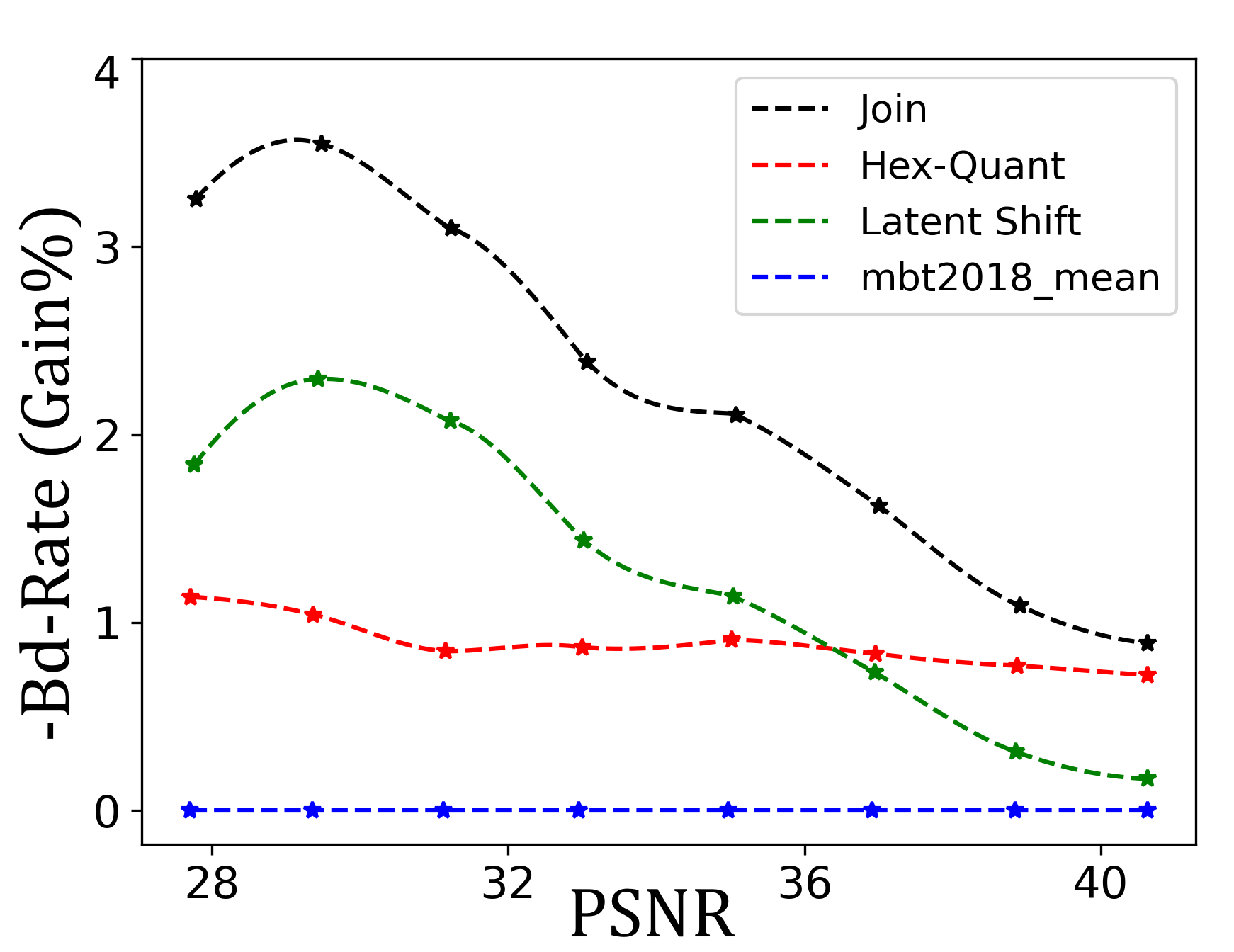}
  }
  \subfigure[] {
    \includegraphics[width=.3\textwidth]{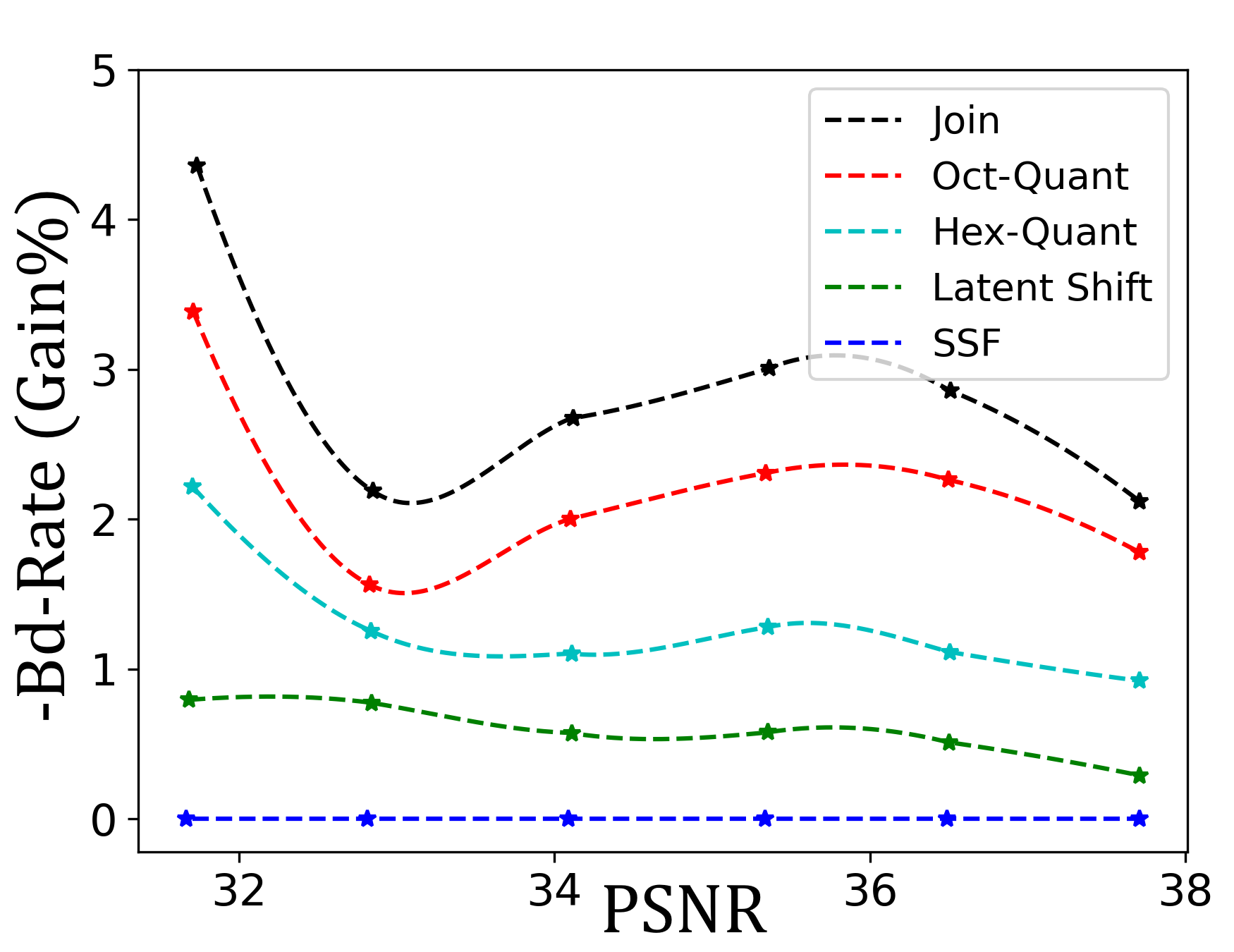}
  }
  \subfigure[] {
    \includegraphics[width=.3\textwidth]{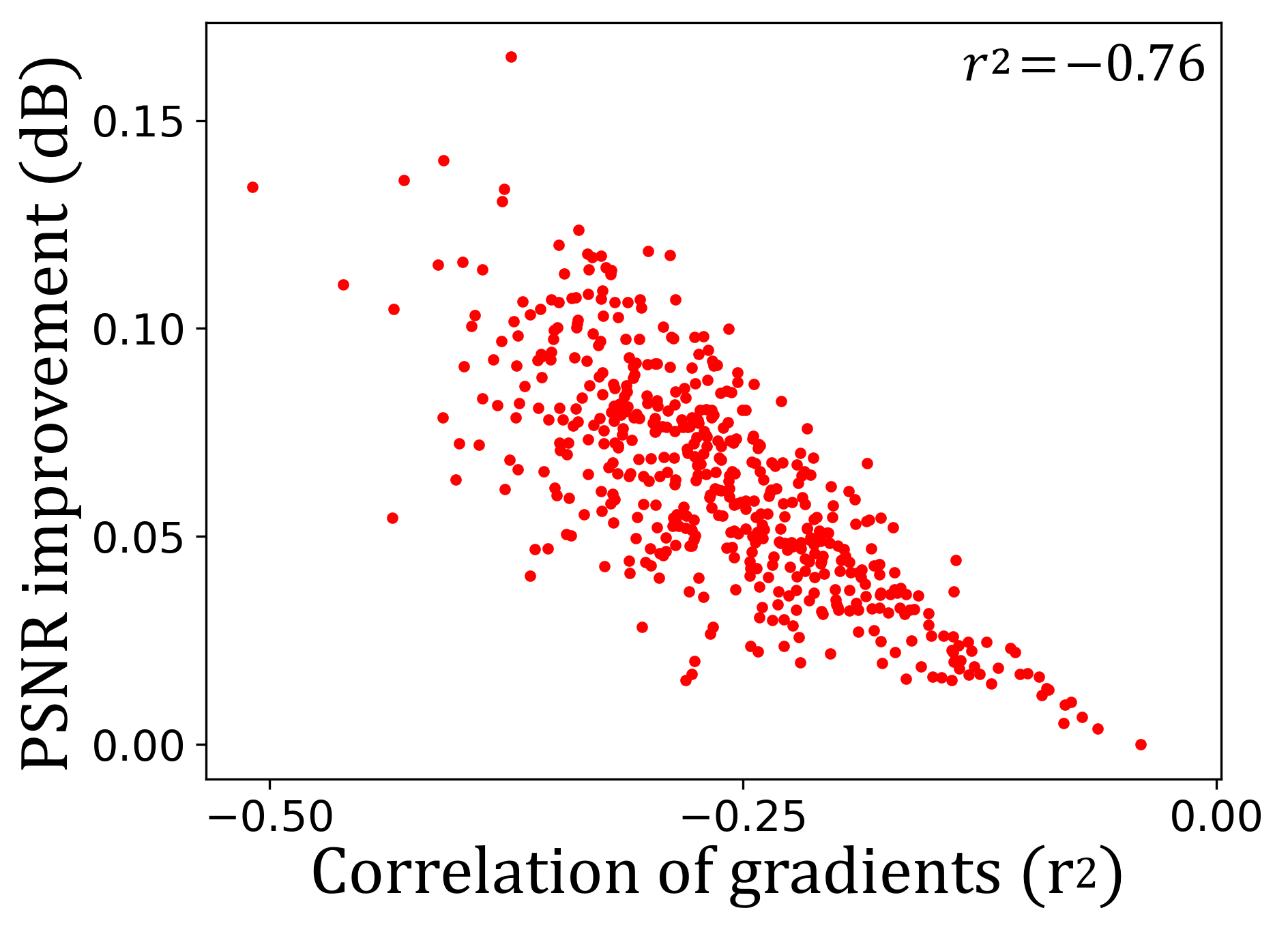}
  }
  \caption{BD-Rates of our proposals from baseline codecs for different quality. a) mbt2018mean image codec on Kodak test set b) SSF video codec on UVG test set. c) Correlation between improvement on reconstruction quality and correlation of gradients on Clic and Kodak datasets.
  }
\label{fig:bdgainkodak}
\medskip
\vspace{-5pt}
\end{figure*}

Figure~\ref{fig:bdgainkodak}a presents the bd-rate (gain in \%) compared to \textbf{mbt2018-mean} for different quality levels on the Kodak dataset. Even though our proposal has a $2.24\%$ gain, the gain on lower quality (thus lower rate) is higher ($3.5\%$). This fact was observed for all codecs for both \textbf{Latent Shift} and \textbf{Hex-Quant} or \textbf{Oct-Quant}. It is because of the fact that VQ is theoretically better than SQ in low bit rate and their performance converges when the bit rate goes infinity. Since the correlation between gradients is stronger in low bit rate rather than high bit rate, \textbf{Latent Shift} performs better in low bit rate.
According to our tests, since the side latent's correlation is weaker, the most important gain of \textbf{Latent Shift} is due to the gradient of the main latent. The analysis of the gain of \textbf{Latent Shift} and its performance against various alternatives is provided in Section~\ref{section:ablationlatent}. 

\begin{table*}[htbp]
\centering
\caption{Average BD-Rate of our proposals for SSF  and DCVC video codecs on UVG dataset and Bunny video in low-delay configuration. \textbf{Join} refers using \textbf{Oct-Quant} and \textbf{Latent Shift}. }\label{tab:ssf}
\begin{tabular}{lcccccccc}
\toprule
 &  \multicolumn{4}{c}{SSF} & \multicolumn{4}{c}{DCVC}\\
 \cline{2-9}

Video & Hex-Quant. & Oct-Quant. & Latent Shift & Join& Hex-Quant. & Oct-Quant. & Latent Shift & Join  \\
\midrule
Beauty & -1.31\% & -1.60\% & -0.13\% & -1.76\% & 
-1.92\%	&-1.97\%	&	-0.10\%	&	-2.02\%	\\
Bosphorus & -1.88\% & -2.29\% & -0.70\% & -3.23\% &
-1.23\%&	-1.48\%&	-0.62\%& -2.05\%\\
Honeybee & -1.75\% & -1.10\%  & -0.88\% & -2.00\% &
-1.80\%&	-2.13\%&	-0.56\%&	-2.65\%\\
Jockey & -1.04\% & -1.48\% & -0.06\% & -1.57\%  &
-1.42\%&	-1.32\%&	-0.11\%&	-1.51\% \\
ReadySteadGo & -1.03\% & -2.00\%& -1.52\% & -3.74\% &
-1.29\%&	-1.30\%&	-1.08\%&	-2.35\%\\
ShakeNDry & -1.38\% & -2.83\% &  -0.35\% & -3.25\% &
-1.55\%&	-1.65\%&	-0.23\%&	-1.92\%\\
YatchRide & -1.03\% & -1.67\% &-0.14\% & -1.87\% &
-1.42\%&	-1.61\%&	-0.18\%&	-1.83\%\\
\midrule
UVG Average & -1.31\% & -1.99\% & -0.60\% & -2.70\% &
-1.52\%&	-1.64\%&	-0.41\%&	-2.05\%\\
\midrule
Bunny & -1.12\% & -1.87\% & -1.31\% & -3.24\% &
-0.90\%&	-1.02\%&	-1.21\%&	-2.25\%
\\
\bottomrule
\end{tabular}
\vspace{-3pt}
\end{table*}
\noindent \textbf{Results on Video Codec:} In Table~ \ref{tab:ssf}, we show the results of the \textbf{SSF} and \textbf{DCVC} video codecs for various sequences. These methods were intended for low-delay mode: frames are coded sequentially using the previously reconstructed frame and an intra frame is inserted every 8 frames (UVG videos are divided into 75 intra periods). I-frame encoding uses the same VAE method as in image codecs. However, each P-frame consists of motion information and residual information encoded by two different VAE models. Since the motion information represents only a fraction of the total bitrate ($3-4\%$), we applied our proposal to the residual coding only. When we apply \textbf{Oct-Quant} and \textbf{Latent Shift} together (\textbf{Join}), it yields an average  bitrate gain of $2-2.7\%$ for the same quality on the UVG dataset. Similarly, the gain at a lower rate is larger (over $4\%$) as demonstrated in Figure~\ref{fig:bdgainkodak}b. 

\noindent \textbf{Gradients correlation:}
The \textbf{Latent Shift}'s gain depends on gradients correlation. To assess this correlation for each individual sample, 
we showed the scatter plot of the individual gains in dB achieved by \textbf{Latent Shift} as well as the actual correlation coefficient between gradients for all test images and all quality levels in Figure \ref{fig:bdgainkodak}c. We observed a strong negative correlation of $-0.76$, independently of the datasets, reconstruction quality or model. 
Since the neural video codec was optimized for the loss of all frames in an intra period (consist of 8 frames), the sum of gradients in \eqref{eq:kkt1} and \eqref{eq:kkt2} is zero in expectation of the frames in an intra period across the datasets. This smooths the sum of gradients even further and decrease the correlations compared to those of image codecs. That explains why the gradient correlation is lower, which leads to a poorer \textbf{Latent Shift} gain for video codecs.

\subsection{Complexity Analysis} 
\label{section:complex}
Since computational complexity without parallelization is a proxy of energy consumption, we performed the test on a single core CPU while discarding multi-thread operations.  We measured the computational time of  \textbf{cheng2020\_attn}, \textbf{mbt2018-mean} on the Kodak dataset and \textbf{SSF} on the bunny dataset. 
The results in Table \ref{tab:complexity} demonstrate the relative encoding/decoding time of our proposed models, namely \textbf{Hex-Quant}, \textbf{Oct-Quant}, and \textbf{Latent Shift}, compared to the baseline models. 
 
\begin{table*}[htbp]
\caption{Encoding and decoding runtime complexity of our proposed methods compare to the baselines.    }
\footnotesize
    \centering    
    \begin{tabular}{lcccccc}
    \toprule
     & \multicolumn{3}{c}{Encoding } & \multicolumn{3}{c}{Decoding } \\
    \cline{2-7}
   Model & Hex Quant &	Oct Quant &	Latent Shift &		Hex Quant &	Oct Quant &	Latent Shift \\
    \hline
    mbt2018-mean   &  +3.0\%&  +5.0\%&  x10.1& +1.8\%  &  +1.8\% &  +0.70\% \\
     cheng2020-attn   &+0.7\% &+1.1\%  & x5.3 & +0.1\% & +0.1\% & +0.06\% \\      
      SSF    &  +2.4\%  & +3.2\%  & x6.0 & +1.9\% & +1.9\%  & +0.70\% \\
\bottomrule
    \end{tabular}
    \label{tab:complexity}
\vspace{-2pt}
\end{table*}

The main source of additional encoding complexity for \textbf{Hex-Quant} and \textbf{Oct-Quant} is the computation of the closest codebook. Naive search can result in up to 20\% overhead complexity for 3-dimensional VQ. However, efficient algorithms \cite{Agrell2002_searchlattices, conway1982} can be used to find the nearest quanta center for hexagons and truncated octahedra. We adapted this solution, thereby reducing the additional encoding time to 3-5\% for the \textbf{mbt2018-mean} model. Although \textbf{cheng2020-attn} and \textbf{mbt2018-mean} have similar absolute extra computational costs, the relative extra cost of \textbf{cheng2020-attn} is lower (1.1-0.7\%) due to the high complexity of the baseline model. For \textbf{SSF}, we ignored the I-frame compression runtime (since it is the same as \textbf{mbt2018-mean}) but measured the average runtime for the encoding and decoding of P-frames, which are done by two VAEs: one for motion information and another for residual information. The results of the relative encoding/decoding time for our \textbf{Hex-Quant}, \textbf{Oct-Quant}, and \textbf{Latent Shift} approach compared to the baseline models can be seen in Table \ref{tab:complexity}. 
Given that the differences in the dimensions of VQ are only reflected in the reshaping of tensors and the number of indices to be retrieved from the dictionary during decoding, the extra decoding complexity  between \textbf{Hex-Quant} and \textbf{Oct-Quant} is negligible. This is almost the same for all three models, as shown in Table \ref{tab:complexity}.

The extra complexity in decoding introduced by the \textbf{Latent Shift} operation is due to the calculation of the entropy gradient with respect to the latents and the shifting operation. Since there is a closed form solution of gradient of gaussian distribution, we do not need to do backward pass to obtain gradient. Instead, we directly calculate the gradient by the parameters of the gaussian distribution. Thus, this extra complexity is negligible, and accounts for less than 1\% of the total decoding time, as shown in Table \ref{tab:complexity}. In contrast, the encoding complexity of \textbf{Latent Shift} is much higher, ranging from 5 to 10 orders of magnitude higher than the baseline encoding time. This is due to the search for a step size out of 8 candidates during encoding, which could be eliminated by using a single universal step size for all images. This solution has been implemented in the \textbf{Latent Shift} extension for traditional codecs, as described in Section \ref{section:ecm80}.

\subsection{Latent Shift versus Alternatives}
\label{section:ablationlatent}
In order to show how the proposed \textbf{Latent Shift} is better than its alternatives, we shifted the latent in a random direction during encoding such that $\mathbf{\hat{y}} \leftarrow \mathbf{\hat{y}}+ \epsilon(\rho_h)$, where $\epsilon(\rho_h) \in \mathbb{R}^{m \times m \times o} \sim \mathcal{N}(0,1)$ and $\rho_h$ is a random seed to be signaled to the decoder. The best random seed in terms of PSNR improvement, should be found at the encoding stage. We generated $1024$ random gradient and encode the best random seed with $10$ bits as extra information. Even though this approach is costly in terms of computational complexity during the encoding stage, (it requires $1024$ times forward passes), we think this is a natural baseline for our proposal, and refer to it as \textbf{Random Shift}. Another alternative would be to use a constant gradient for all latents in a given image. Thus, at encoding, we need to test a large set of values and assume all latents may be shifted by this amount such that $\mathbf{\hat{y}} \leftarrow \mathbf{\hat{y}}+ \rho_h$ where $\rho_h \in \mathbb{R}$. The best value of $\rho_h$ should be signaled to the decoder with $10$ bit extra cost. We refer to this as the \textbf{Scalar Shift} approach. 

Our final alternative does not use the true gradient with respect to entropy directly, but instead relies on an approximation. Specifically, we make the assumption that the rate of change of any latent is independent and increases linearly with the distance from the center  as $R(\mathbf{\hat{y}})=a|\mathbf{\hat{y}}-{\bm \mu}|+b$ where $a,b \in \mathbb{R}^+$. 
Under this assumption, the magnitude of the gradient will be constant, but the direction will always point away from the center. Thus, this approach shifts the latent in the opposite direction to the distribution's center such that $\mathbf{\hat{y}} \leftarrow \mathbf{\hat{y}}- \rho_h\operatorname{sign}(\mathbf{\hat{y}}-{\bm \mu})$. In hyperprior entropy models, the latent is assumed to follow a Gaussian distribution, the latent's entropy thus gets smaller if it moves towards the center of the distribution. By shifting the latent in the opposite direction, the entropy increases. Since the factorized entropy model does not use a Gaussian distribution, we shifted the latent to the opposite direction of the zero center means assuming ${\bm \mu}=0$ in the factorized entropy. The best value of $\rho_h$ should be signaled to the decoder with a $10$ bit extra cost. We refer to this approach as \textbf{Sign Shift}.
\vspace{-2pt}
\begin{table*}[htbp]
\centering
\caption{Average BD-PSNR of \textbf{Latent Shift} and some alternatives.}\label{tab:latentshiftab}
\footnotesize
\begin{tabular}{lcccc}
\toprule
Baseline Codec&  Random Shift & Scalar Shift & Sign Shift & Latent Shift\\
\midrule
bmshj2018-factorized &  0.0002 dB & 0.0036 dB & 0.0151 dB & 0.0297 dB \\
mbt2018-mean &  0.0006 dB& 0.0007 dB & 0.0339 dB& 0.0705 dB\\
\bottomrule
\end{tabular}
\vspace{-2pt}
\end{table*}

In Table~ \ref{tab:latentshiftab}, we present the results for \textbf{bmshj2018-factorized} (lowest correlation between gradients) and \textbf{mbt2018-mean} (highest correlation between gradients). All alternative approaches require $10$ extra bits, however we ignored these bits in our comparison, for which we used  the BD-PSNR metric defined in \cite{bjontegaard2001calculation}. These results show that our proposal is significantly better than all the alternatives considered in this study. The closest one, \textbf{Sign Shift}, reaches half the performance of \textbf{Latent Shift}. The random alternative could not improve the baseline significantly, even though their encoder is almost $1000$ times more computationally demanding.  

\subsection{Enhancing Traditional Codecs with Latent Shift}

\label{section:ecm80}

Despite traditional codecs not using gradient-based optimization methods and instead relying on discrete search algorithms within a restricted search space for minimizing RD costs, we can still argue that the KKT condition exists. This is because the discrete search algorithms used by traditional codecs find the optimum transform coefficient of residuals (known as latent variables in the neural compression literature) for both rate and distortion objectives. 

In order to implement \textbf{Latent Shift} on the current traditional Sota codec \textbf{ECM} described in \cite{ecm80}, we need an entropy model to calculate its gradient w.r.t the transform coefficients. However, since \textbf{ECM} uses a bit-level adaptive entropy model (CABAC), it is not possible to obtain a closed-form solution for the gradient. This may introduce additional complexity if we obtain the gradient by using finite difference methods. To avoid this complexity, we assume each coefficient are independent, and the rate (entropy of the coefficients) linearly increases by logarithm of the absolute value of the coefficient and an offset $b$ such as $R(|\mathbf{\hat{y}}|) = a \log(|\mathbf{\hat{y}}|) +b, |\mathbf{\hat{y}}| \geq 1$. This assumption leads us to write gradient of the entropy by 
$\nabla_{|\mathbf{\hat{y}}|}R(|\mathbf{\hat{y}}|)=\frac{a}{|\mathbf{\hat{y}}|}$. When we use universal optimal step size $\rho^*$, \textbf{Latent Shift} can be seen as shifting the coefficients in de-quantization step by $|\mathbf{\hat{y}}| \leftarrow |\mathbf{\hat{y}}|+\frac{\alpha}{|\mathbf{\hat{y}}|}$ where $\alpha=a\rho^*$ is a single parameter of the method. We fine-tuned $\alpha$ over train set and priorly hard coded into both the sender and receiver. More details can be found in JVET document in \cite{jvetae0125}.

\vspace{-2pt}
\begin{table*}[htbp]
\caption{Bd-rate of \textbf{Latent Shift} method for Y, U and V channel and relative encoding and decoding time percentage compare to ECM-9.0 on All Intra and Random Access mode under Common Test Conditions. 
    }
    \centering    
    \begin{tabular}{lcccccccccc}
    \toprule
       & \multicolumn{5}{c}{All Intra (Image compression) } & \multicolumn{5}{c}{Random Access (Video Compression) } \\   
    \cline{2-11}
   Sequences & Y &	U &	V & EncT & DecT & Y &	U &	V & EncT & DecT \\
    \hline    
Class A1 & -0.10\% & -0.16\% & -0.23\% & 101\% & 100\% & -0.06\% & -0.39\% & -0.09\% & 99\% & 98\%\\
Class A2 & -0.06\% & -0.16\% & -0.04\% & 99\% & 98\% & -0.04\% & -0.09\% & -0.25\% & 95\% & 98\%\\
Class B & -0.12\% & -0.04\% & -0.14\% & 101\% & 98\% & -0.07\% & -0.13\% & -0.19\% & 103\% & 102\%\\
Class C & -0.08\% & -0.19\% & -0.12\% & 100\% & 102\% & -0.05\% & -0.46\% & -0.20\% & 103\% & 102\%\\
Class E & -0.13\% & -0.07\% & 0.01\% & 100\% & 101\% \\
Overall & -0.10\% & -0.12\% & -0.11\% & 100\% & 100\% & -0.06\% & -0.26\% & -0.18\% & 100\% & 100\%\\
\bottomrule
    \end{tabular}  
    \label{tab:ecmresult}
\vspace{-5pt}
\end{table*}

The results presented in Table \ref{tab:ecmresult} were obtained under the Common Test Conditions (CTC) \cite{ecm80ctc}. According to the results, \textbf{Latent Shift} managed to save $0.1\%$ on the Luma channel and more than that on the chroma channels in all intra-mode settings where each frame is encoded separately. In Random Access mode, which is the most effective video compression setting, the Luma results were a little lower while chroma results were better. Thanks to the universal step size and the approximation of the gradient, our proposal has no impact on the encoding and decoding times, as reported in Table \ref{tab:ecmresult}. This proposal is adopted to ECM-10.0 by JVET. The contributed code can be found on reference software. \footnote{https://vcgit.hhi.fraunhofer.de/ecm/ECM/-/blob/master/source/Lib /CommonLib/DepQuant.cpp}

\subsection{Latent Shift after Fine-tuning Solutions }
To demonstrate the orthogonality between \textbf{Latent Shift} and an encoder-side fine-tuning solution proposed in \cite{Campos_2019_CVPR_Workshops}, we conducted tests on two baseline models: one with fine-tuning and one without. In the fine-tuning approach, the latents were fine-tuned for 1000 iterations during encoding, which incurred 1000 forward passes and 1000 backward gradient calculations, thus resulting in around 4000 times more complexity without parallelization. In contrast, our \textbf{Latent Shift} incurred negligible extra-encoding time compared to fine-tuning solutions whose complexity can be found in Table \ref{tab:finetune}.

An important result can be observed when applying \textbf{Latent Shift} after fine-tuning solutions, as shown in Table \ref{tab:finetune}. Fine-tuning solutions minimize the loss without averaging images in the training set, which strengthens the KKT conditions and increases correlations between gradients. For instance, the average correlation of gradients for the Kodak dataset becomes $-0.2212$ after fine-tuning, compared to $-0.1805$ in the \textbf{mbt2018-mean} codec. The improved correlations lead to improved performance, as shown by the fact that the fine-tuning solution only achieves a rate saving of $-5.77\%$, while combining it with \textbf{Latent Shift} increases the rate saving to $-7.47\%$ for the Kodak dataset.

\begin{table*}[htbp]
\centering
\caption{\textbf{Latent Shift} performance over Fine-tuning solutions}
\label{tab:finetune}
\footnotesize
\begin{tabular}{lllcccc}
\toprule
Model&	EncT&	DecT&	Bd-Rate (Kodak)	&Corr. (Kodak)&	Bd-Rate (Clic)&	Corr.  (Clic) \\
\midrule
bmshj2018 baseline  & x1  & +0.0\% & 0.0\% &  & 0.0\% &   \\
Only \textbf{Latent Shift} &  x12&	+0.7\%&	-0.49\% &	-0.108&	-0.69\%	&-0.0952  \\
Only Fine-Tuning  & x4800&	+0.0\%	&-6.52\%	&	&-6.73\%	&  \\
Fine-Tuning + \textbf{Latent Shift} &x4812&	+0.7\%&	-7.88\%&	-0.165&	-8.06\%&	-0.1578  \\
\midrule
mbt2018-mean baseline & x1  & +0.0\%  &	0.0\%	&  &	0.0\%&  \\
Only \textbf{Latent Shift} & x10.1 &	+0.7\%	&-1.27\%	&-0.1805	&-1.21\% &	-0.1465   \\
Only Fine-Tuning  & x4380 &	+0.0\% &	-5.77\%	& 
 &	-5.53\%	&  \\
Fine-Tuning + \textbf{Latent Shift} & x4390 &	+0.7\%&	-7.47\% &	-0.2212	&-7.16\% &	-0.1796 \\
\bottomrule
\end{tabular}
\vspace{-5pt}
\end{table*}

\section{Conclusion}
\label{sec:conc}
In this work, we proposed two orthogonal methods for further improving the latent representation of generic compressive variational auto-encoders (VAE). We initially exploited the remaining redundancy in the latents during the quantization stage. To do so, we demonstrated that a uniform VQ method improves VAEs, when trained using uniform scalar quantization. Secondly, we used the correlation between the entropy gradient and the reconstruction error to improve latent representation. The combination of these two methods improves latent representation, and brings significant gains to several state-of-the-art compressive auto-encoders, without any need for retraining.
Based on these results, several improvements can be foreseen. First, the correlation between the gradients depends on the training set according to the definition of the KKT conditions. However, even though different models use the same training set and training procedure, their gradients' correlation coefficient may be different. It would thus be interesting to explore the connection between certain type of model architecture and the correlation between the gradients.
Secondly, the uniform VQ process was applied without retraining. Even though VQ quantization error is lower on average, maximum quantization error is sometimes higher. Since the decoder block is not aware of these higher individual errors during  training, the model becomes sub-optimal. This could be solved by re-training the decoder block or the entire model using a continuous relaxation of the uniform VQ grid. Last but not least, even though it is straight forward to extent our VQ proposal to higher than $3$ dimensions, the number of unique codes exponentially increases by the dimension which makes arithmetic encoder with current bit resolution impossible to represent the probabilities. Thus, some tricks such as removing some grids can be done for higher dimensional VQ as a future work.

\appendices
\section{Proof of Theorem 1}
\label{section:pth1}
\begin{proof}
The scalar quantization map can be see as $n+1$ border values with $b_0<b_1< \dots <b_n$ and $n$ quantization center with $c_1<c_2< \dots < c_n$. Any latent $y$ with  $b_{i-1}\leq y < b_i$ should be quantized to point $c_i$. The quantization map can be represented by an ordered set consisting of all borders and centers such as $\mathbb{M}=\{b_0<c_1<b_1<c_2<b_2< \dots <c_n<b_n\}$ where $|\mathbb{M}|=2n+1$. When the quantization map is non-uniform, the difference between consecutive elements in the set are not necessarily equal, i.e. $\exists (i,j),\mathbb{M}_i-\mathbb{M}_{i-1} \ne \mathbb{M}_j-\mathbb{M}_{j-1}$. The nearest integer quantization map can be defined as $\mathbb{M}^{(u)}=\{0.5,1,1.5,2,2.5, \dots n,n+0.5\}$ or simply $\mathbb{M}^{(u)}_i=0.5i$ thus,$\forall i, \mathbb{M}^{(u)}_i-\mathbb{M}^{(u)}_{i-1}=0.5$. Let's assume that any arbitrary $\mathbb{M}$ is the optimal scalar quantization map of a neural codec's latent $y$ obtained by $y=g_a(x)$. Since $\mathbb{M}$ and $\mathbb{M}^{(u)}$ are both monotonic increasing set, there exists a bijective function $f(.)$ that maps $\forall i, \mathbb{M}_i$ to $\mathbb{M}^{(u)}_i$, and $f^{-1}(.)$ maps $\mathbb{M}^{(u)}_i$ to $\mathbb{M}_i$, $\forall i$. 
According to the universality theorem \cite{hornik1989multilayer}, function $f(.)$ can be implemented by a multi layer neural network.

Two codecs' performances are equal if and only if the entropy of their latents are  equal and their reconstructions are the same. First, we start by showing that their entropies are the same by showing that the corresponding center's PMFs are equal in both spaces, i.e. $\forall i, P(i)=P^{(u)}(i)$. We can write the $i$-th quantization center's PMF $P(i)=\int_{b_{i-1}}^{b_i}p(y)dy$ where $y$ is a point in $g_a$'s output space. Any point $y$ can be transformed into a new space by $z=f(y)$. We can write the $i$-th quantization center's PMF as $P^{(u)}(i)=\int_{i-0.5}^{i+0.5}p(z)dz$ in this space. We can rewrite it as $P^{(u)}(i)=\int_{f^{-1}(i-0.5)}^{f^{-1}(i+0.5)}p(f^{-1}(z))df^{-1}(z)$. Since $f^{-1}(i+0.5)=b_i$, $f^{-1}(i-0.5)=b_{i-1}$ and $f^{-1}(z)=y$, we can write $\forall i,P^{(u)}(i)=\int_{b_{i-1}}^{b_i}p(y)dy=P(i)$.

The output of a deep decoder is $g_s(c_i)$ if latent $y=g_a(x)$ meets $b_{i-1}\leq y < b_i$. Any latent $b_{i-1}\leq y < b_i$ is mapped to $f(b_{i-1})\leq f(y) < f(b_i)$ thus $i-0.5\leq z < i+0.5$. The  latent $z$ which lies between $i-0.5$ to $i+0.5$, can be quantized to $i$ in the new space. Since the decoder applies $g_s(f^{-1}(z))$, its output should be $g_s(f^{-1}(i))$. Since $f^{-1}(i)=c_i$, it produces $g_s(c_i)$. 
\end{proof}

\section{Advantage of Space Tessellation Grid on Quantization}
\label{section:hexQuant}
When the source has a uniform distribution, quantization using a truncated octahedron produces better RD performance compared to regular hexagonal grids, and a regular hexagonal grid leads to better RD performance than uniform SQ grids (nearest integer rounding). In order to show the superiority of a method over another in terms of RD performance, comparing the reconstruction error at an equal bit-rate sufficient. Since the equal volume grids have the same probability under uniform distribution, the rate is equal for all three cases. Since the distributions are identical, we just need to compute the mean squared error for each types of grid in a specific position, for example the origin for 1D, 2D and 3D cases respectively. 

The MSE of a uniform SQ grid can be written as integral of the square error normalized by the grid size: 
\begin{equation}
   \label{eq:sqint}
   MSE^{(s)}(u)=\frac{1}{u}\int_{-u/2}^{u/2} x^2dx.
\end{equation}

This produces $MSE^{(s)}(u)=u^2/12 \approx 0.0833u^2$. 
\begin{figure*}[htbp]
  \centering
  \subfigure[] {
    \includegraphics[width=.4\textwidth]{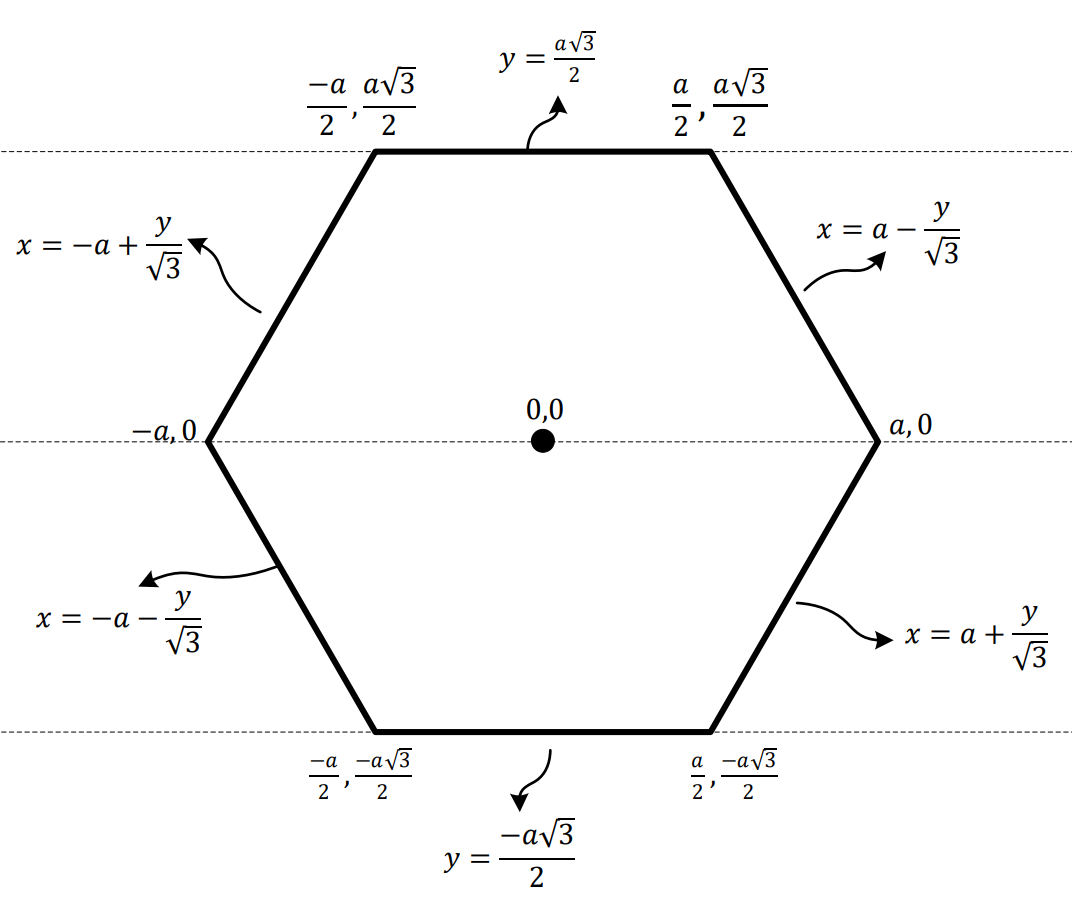}
  }
 \subfigure[] {
 \includegraphics[width=.35\textwidth]{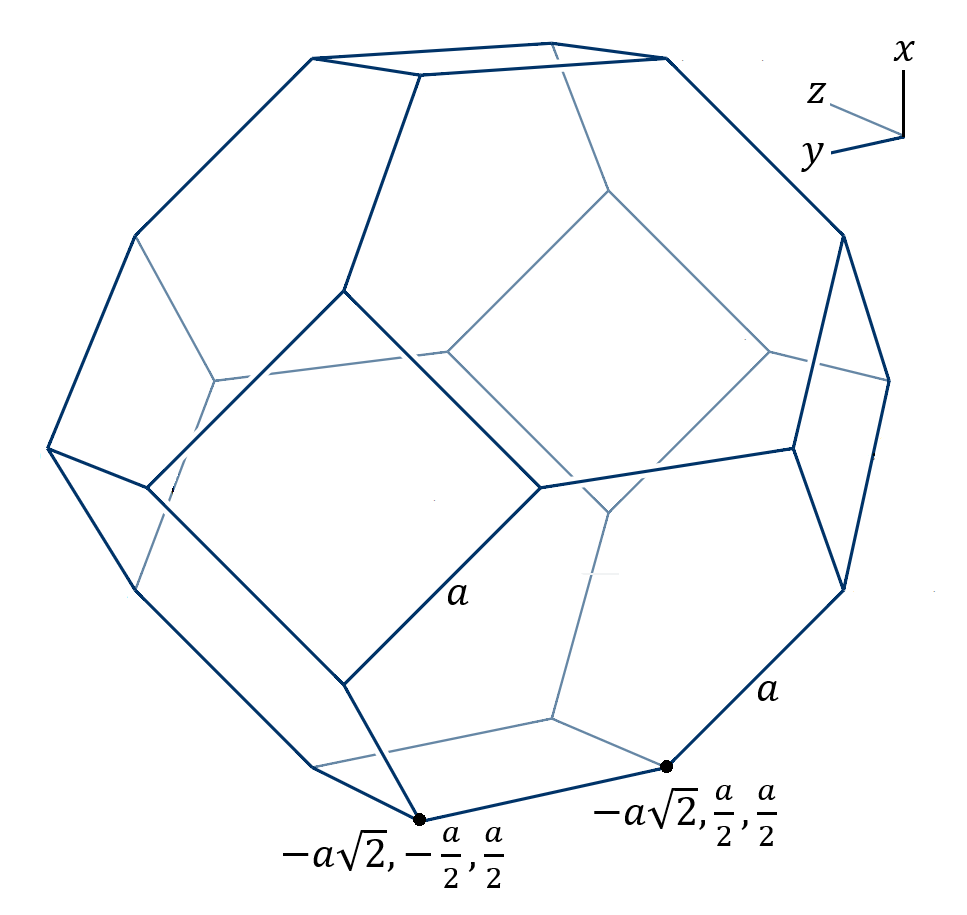}
  }

\caption{ a) Regular hexagon and b) Truncated octahedron located at origin. }
\label{fig:hex}
\end{figure*}

\noindent \textbf{Hexagonal grid case:}
For hexagonal grid, we need to double integrate over the hexagonal domain. Figure \ref{fig:hex}a shows an hexagon with a side a length $a$ located at the origin. We also show each side's functions in a 2D space. We divide the hexagon into two parts for positive and negative $y$ and calculate the analytic integral for these two region separately. We then normalize the sum of squares by the area of the hexagon, which is $3\sqrt{3}a^2/2$.

\begin{multline}
   \label{eq:sqint2}
   MSE^{(h)}(a)=\frac{2}{3\sqrt{3}a^2}\left(\int_{0}^{a\frac{\sqrt{3}}{2}} \int_{-a+\frac{y}{\sqrt{3}}}^{a-\frac{y}{\sqrt{3}}} \frac{x^2+y^2}{2}dxdy + \right. \\
   \left. \int_{-a\frac{\sqrt{3}}{2}}^{0} \int_{-a-\frac{y}{\sqrt{3}}}^{a+\frac{y}{\sqrt{3}}}\frac{ x^2+y^2}{2} dxdy\right). 
\end{multline}

We first integrate the positive part of the integral over $x$ between  $-a+\frac{y}{\sqrt{3}}$ and $a-\frac{y}{\sqrt{3}}$, followed by the integral over $y$ between $0$ and $a\frac{\sqrt{3}}{2}$. The integral for the positive region is then $\frac{5\sqrt{3}a^4}{32}$. The same can be done for the negative region, but since the distribution is uniform, it produces the exact same result, we thus conclude that $MSE^{(h)}(a)=\frac{5a^2}{24}$. In order to obtain the same volume for the grid, the hexagon's area should be $u^2$. Since the hexagon's area is $3\sqrt{3}a^2/2$ for a side length of $a$, we find that $a=\frac{\sqrt{2}u}{\sqrt{3\sqrt{3}}}$. Thus $MSE^{(h)}(u)=\frac{5\sqrt{3}u^2}{108} \approx 0.0801u^2$.

\noindent \textbf{The truncated octahedron case:} For the MSE of the truncated octahedral grid, we need to integrate the mean squared error of each of the $3$ dimensions over the truncated octahedral domain. In figure \ref{fig:hex}b, a truncated octahedral with a side length of $a$ is shown. The solution can be obtained by integrating one octant, multiplying it by $8$ and normalizing by its volume which is $8\sqrt{2}a^3$ as follows:

\begin{multline}
\scriptstyle
   \label{eq:sqint3}
   MSE^{(o)}(a)=\frac{8}{8\sqrt{2}a^3}\int_{0}^{\sqrt{2}a} \int_{0}^{min(\sqrt{2}a,3\sqrt{2}a/2-x)}\\ \int_{0}^{min(\sqrt{2}a,3\sqrt{2}a/2-x-y)}
   \left( \frac{x^2+y^2+z^2}{3} \right)dzdydx. 
\end{multline}
First, we integrate over $z$, $y$ and then $x$, which produces solution $MSE^{(o)}(a)=\frac{19}{48}a^2$. To compare with the same grid volume, the truncated octahedron should have a volume of $u^3$. Since the volume is $8\sqrt{2}a^3$ when one side length is $a$, we find that $a=\frac{u}{(8\sqrt{2})^{1/3}}$, thus $MSE^{(o)}(u)=\frac{19}{48(8\sqrt{2})^{2/3}}u^2  \approx 0.0785u^2$.  

As a result, $MSE^{(s)}\approx0.0833u^2$, $MSE^{(h)}\approx0.0801u^2$ and $MSE^{(o)}\approx0.0785u^2$, where the volume of the grid is $u$. We can thus conclude that $\forall u \in \mathbb{R}^+, MSE^{(o)}(u)< MSE^{(h)}(u) < MSE^{(s)}(u)$.

\section{Proof of Theorem 2 and Corollary 2.1}
\subsection{Theorem 2}
\label{section:pth2}

\begin{proof}
Since $\mathbf{\hat{z}}$ is a function of $\phi,\Phi$ and $\mathbf{x}$, we can define a function for the first objective of \eqref{eq:loss} by $\mathcal{L}_1(\mathbf{x},\phi,\Phi,\Psi):=-\log(p_f(\mathbf{\hat{z}};\Psi))$, and $\mathbf{\hat{y}}$ is a function of $\phi$ and $\mathbf{x}$ we can define a function for the second objective of \eqref{eq:loss} by $\mathcal{L}_2(\mathbf{x},\phi,\Phi,\Theta):=-\log(p_h(\mathbf{\hat{y}};\mathbf{\hat{z}},\Theta))$. For the third objective of \eqref{eq:loss}, we can define a function by $\mathcal{L}_3(\mathbf{x},\phi,\mathbf{\theta}):=d(\mathbf{x},g_s(\mathbf{\hat{y}};\mathbf{\theta}))$. When considering the coefficient of the objectives as defined by $\alpha_1:=1/(2+\lambda)$, $\alpha_2:=1/(2+\lambda)$, and $\alpha_3:=\lambda/(2+\lambda)$, we can write \eqref{eq:loss} as follows.

\begin{equation}
\begin{aligned}
   \label{eq:loss_th2}
   \phi^*,\theta^*,\Phi^*,\Theta^* \Psi^*=\argmin_{\phi,\theta,\Phi,\Theta, \Psi} \left ( \mathbb{E}_{\mathbf{x}\sim p_x} \left[ \alpha_1\mathcal{L}_1(\mathbf{x},\phi,\Phi,\Psi)+ \right. \right. \\
   \left. \left. \alpha_2\mathcal{L}_2(\mathbf{x},\phi,\Phi,\Theta)+\alpha_3\mathcal{L}_3(\mathbf{x},\phi,\mathbf{\theta})    \right]     \right).
\end{aligned}
\end{equation}

Since $\lambda > 0, \forall i, \alpha_i > 0$ and $\sum_i \alpha_i = 1$, the set of $\alpha_i$s corresponds to coefficients set and $\mathcal{L}_i(.)$s corresponds to objectives, \eqref{eq:loss_th2} shows an unconstrained multi-objective optimization problem. This problem has five sets of variables to be optimized and gradients w.r.t each variable set should meet the KKT conditions. Since $\mathcal{L}_3(\mathbf{x},\phi,\mathbf{\theta})$ does not depend on $\Phi$, thus, $\nabla_{\Phi}\mathcal{L}_3(\mathbf{x},\phi,\mathbf{\theta})=0$, so  we can write the KKT conditions w.r.t $\Phi$ as follows:

\begin{equation}
   \label{eq:loss_th2_1}
   \notag
     \mathbb{E}_{\mathbf{x}\sim p_x} \left[ \alpha_1\nabla_{\Phi}\mathcal{L}_1(\mathbf{x},\phi,\Phi,\Psi)+\alpha_2\nabla_{\Phi}\mathcal{L}_2(\mathbf{x},\phi,\Phi,\Theta))    \right]=0.
\end{equation}

Since $\alpha_1=\alpha_2$, we obtained the first condition in \eqref{eq:kkt1} by simply replacing $\mathcal{L}_1(.)$ and $\mathcal{L}_2(.)$ with their definitions. Similarly,  we can write the KKT conditions w.r.t $\phi$ as follows:

\begin{equation}
\begin{aligned}
   \label{eq:loss_th2_3}
   \notag
     \mathbb{E}_{\mathbf{x}\sim p_x} \left[ \alpha_1\nabla_{\phi}\mathcal{L}_1(\mathbf{x},\phi,\Phi,\Psi)+\alpha_2\nabla_{\phi}\mathcal{L}_2(\mathbf{x},\phi,\Phi,\Theta))  + \right.
     \\
     \left. \alpha_3\nabla_{\phi}\mathcal{L}_3(\mathbf{x},\phi,\mathbf{\theta})  \right]=0.
\end{aligned}
\end{equation}

This is equivalent to \eqref{eq:kkt2} when $\alpha_i$s and $\mathcal{L}_i(.)$s
are replaced with their definitions.

\end{proof}

\subsection{Corollary 2.1}
\label{section:pcol2}

\begin{proof}
When we remove the expectation term from \eqref{eq:kkt1}, we get the followings:

\begin{equation}
\label{eq:col_pr1}
    \notag
    \nabla_{\Phi}\log(p_f(\mathbf{\hat{z}};\Psi)) =- \nabla_{\Phi}\log(p_h(\mathbf{\hat{y}};\mathbf{\hat{z}},\Theta)).     
\end{equation}

Since $\mathbf{\hat{z}}$ and $\Phi$ are dependent ($\mathbf{\hat{z}}=Q(h_a(\mathbf{y};\Phi)$) partial derivations do not equal zero. Thus, if we multiply both sides with Jacobian matrix $\mathbb{J}^{(\Phi,\mathbf{\hat{z}})}$ where $\mathbb{J}^{(\Phi,\mathbf{\hat{z}})}_{i,j}=\pdv{\Phi_j}{\mathbf{\hat{z}}_i}$, we change the gradient variable from $\Phi$ to $\mathbf{\hat{z}}$ and reach \eqref{eq:col1}.   

We can repeat the same steps for \eqref{eq:col2}. When we remove the expectation term from \eqref{eq:kkt2}, we get the following:

\begin{equation}
\begin{aligned}
\label{eq:col_pr2}
\notag
    \nabla_{\phi}\left [-\log(p_f(\mathbf{\hat{z}};\Psi))-\log(p_h(\mathbf{\hat{y}};\mathbf{\hat{z}},\Theta)) \right] =  \\
    -\lambda  \nabla_{\phi}d(\mathbf{x},g_s(\mathbf{\hat{y}};\mathbf{\theta})). 
\end{aligned}    
\end{equation}

Since $\mathbf{\hat{y}}$ and $\phi$ are dependent ($\mathbf{\hat{y}}=Q(g_a(\mathbf{x};\phi)$) partial derivatives do not equal zero. Thus, if we multiply both sides with Jacobian matrix $\mathbb{J}^{(\phi,\mathbf{\hat{y}})}$ where $\mathbb{J}^{(\phi,\mathbf{\hat{y}})}_{i,j}=\pdv{\phi_j}{\mathbf{\hat{y}}_i}$, we change the gradient variable from $\phi$ to $\mathbf{\hat{y}}$ and reach \eqref{eq:col2}.  

\end{proof}
\section{Gain Analysis of Latent Shift}
\label{sec:ablation}
\begin{table*}[!htbp]
\centering
\caption{Upper limits of the gradient based latent shifting on \textbf{mbt2018-mean} codec.}\label{tab:gainlatent}
\footnotesize
\begin{tabular}{lcccc}
\toprule
{} &  Only Side Shift & Only Main Shift & Only Main Shift & Main \& Side Shift \\
{}& (BD-Rate) & (BD-Psnr) & (BD-Rate) & (BD-Rate)\\ 
\midrule
\textbf{True Gradients} &  -1.011\% &  1.3972 dB & -25.139\% & -26.150\% \\
\textbf{Latent Shift} &  -0.031\% & 0.0705 dB & -1.270\% & -1.301\% \\
\bottomrule
\end{tabular}
\end{table*}

\begin{figure}
\begin{center}
    \includegraphics[width=.48\textwidth]{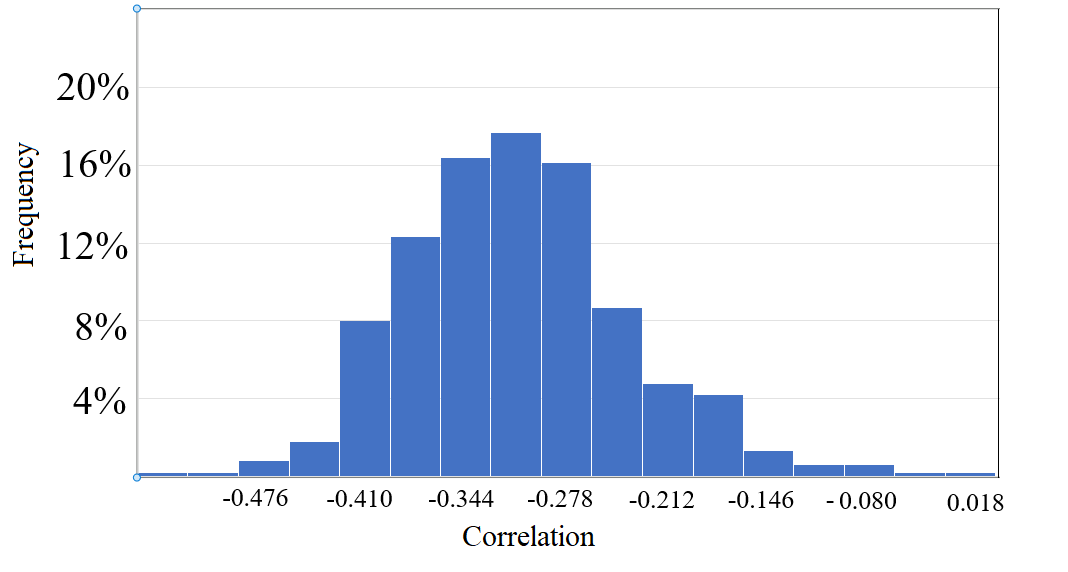}
\end{center}
\caption{Histogram of correlation between gradients wrt main latents. The data is taken with \textbf{mbt2018-mean} image codec on Kodak and Clic dataset.}
\label{fig:corr_hist}
\end{figure}

To understand the upper limits of gradient-based latent shifting, we used true gradients instead of proxy ones and measured the performance. We simply shifted the side latents by $\mathbf{\hat{z}} \leftarrow \mathbf{\hat{z}}+\rho_f^*\nabla_{\mathbf{\hat{z}}}(-log(p_h(\mathbf{\hat{y}};\mathbf{\hat{z}},\Theta)))$ after decoding $\mathbf{\hat{z}}$. Later, we shift the main latent by $\mathbf{\hat{y}} \leftarrow \mathbf{\hat{y}}+\rho_h^*\nabla_{\mathbf{\hat{y}}}(d(\mathbf{x},g_s(\mathbf{\hat{y}};\mathbf{\theta})))$ after decoding $\mathbf{\hat{y}}$. We can see this hypothetical case as if the correlations were $-1$. This case is mentioned by \textbf{True Gradients} in the ~results in Table~\ref{tab:gainlatent} while our proposal is \textbf{Latent Shift}. The results are obtained with the \textbf{mbt2018-mean} image codec on the Kodak dataset. 

According to these results, we can see that even if the side latent's gradients were perfectly correlated, our maximum gain would be around $1\%$. Since the gradient correlation w.r.t. the side latent is weak ( $r^2 \approx -0.07$), our gain is negligible. As a consequence, in practice, shifting side latent may not be ignored. On the other hand, the true gradients of main latents increase PSNR by $1.40$dB on average, which is equivalent to saving around $25\%$ of the bitstream. Our proposal could increase the PSNR by $0.07$dB on average which is equivalent to saving $1.27\%$ of the bitstream for the same quality thanks to the existing correlation between gradients w.r.t. the main latents as shown in Figure~\ref{fig:corr_hist}. The gain is of course smaller than the upper limit, but significant still. Since keeping these gradients is costly (nearly as costly as saving the image itself), searching for a more effective way of using those gradients makes sense.

\section{Numeric PMF Calculation for Hexagonal Domain}
\label{section:alg1}
The closed-form solution to the integrals of known multidimensional densities over any domain includes very specific solutions, and generally no tractable form \cite{savaux2020computation}. This also applies to the Gaussian distribution of dimension $2$ on the regular hexagonal domain. There is therefore no closed form solution to \eqref{eq:hexpmf}, where the probabilities are independent Gaussian on the regular hexagonal domain. However, we found the solution using the combination of both analytical and numerical integration as detailed below.

The integral centered around $(G_x,G_y)$ of hexagonal domain $G$ for the independent Gaussian distribution is given by 
\begin{equation}
   \label{eq:intgg}
   P(G_x,G_y)=\int_{G} N(x, \mu_1, \sigma_1).N(y, \mu_2, \sigma_2).dx.dy.
\end{equation}
For the sake of simplicity, let our grid be located in the center $(G_x,G_y)=(0,0)$ and $a$ be one side length of the hexagon. Then, the integral is defined as 
\begin{multline}
   \label{eq:intcgg}
   \scriptstyle
   P(0,0)=\int_{0}^{a\frac{\sqrt{3}}{2}} \int_{-a+\frac{y}{\sqrt{3}}}^{a-\frac{y}{\sqrt{3}}} N(x, \mu_1, \sigma_1).N(y, \mu_2, \sigma_2).dx.dy + \\ 
   \scriptstyle
   \int_{-a\frac{\sqrt{3}}{2}}^{0} \int_{-a-\frac{y}{\sqrt{3}}}^{a+\frac{y}{\sqrt{3}}}N(x, \mu_1, \sigma_1).N(y, \mu_2, \sigma_2) .dx.dy, 
\end{multline}
where the hexagon is divided into lower and upper half parts, and we add up the two integrals. It is noted that the integral does not admit a closed form solution, but the inner integral does and the outer integral does not have the analytical solution. The solution of the inner integral in the first part is given by
\begin{equation}
   \label{eq:intinner}
   \scriptstyle
    \int_{-a+\frac{y}{\sqrt{3}}}^{a-\frac{y}{\sqrt{3}}} N(x, \mu_1, \sigma_1).dx = \frac{1}{2}erf\left(\sqrt{2}\frac{(-\mu_1 + x)}{2\sigma_1}\right)\bigg|_{x=-a-\frac{y}{\sqrt{3}}}^{x=a-\frac{y}{\sqrt{3}}}.
\end{equation}
By substituting into \eqref{eq:intcgg}, we obtain:
\begin{multline}
   \label{eq:intfin}
   \scriptstyle
   \int_{0}^{a\frac{\sqrt{3}}{2}} \int_{-a+\frac{y}{\sqrt{3}}}^{a-\frac{y}{\sqrt{3}}} N(x, \mu_1, \sigma_1).N(y, \mu_2, \sigma_2).dx.dy  \\
   \scriptstyle
   = \int_{0}^{a\frac{\sqrt{3}}{2}} \frac{1}{2}erf\left(\sqrt{2}\frac{(-\mu_1 + x)}{2\sigma_1}\right)\bigg|_{x=-a-\frac{y}{\sqrt{3}}}^{x=a-\frac{y}{\sqrt{3}}}
   N(y, \mu_2, \sigma_2).dy.
\end{multline}
This is finally solved using numerical integration\footnote{https://github.com/esa/torchquad}. We can also similarly obtain the integral of the second part. 

\bibliography{main}
\bibliographystyle{IEEEtran}
\begin{IEEEbiography}[{\includegraphics[width=1in,height=1.25in,clip,keepaspectratio]{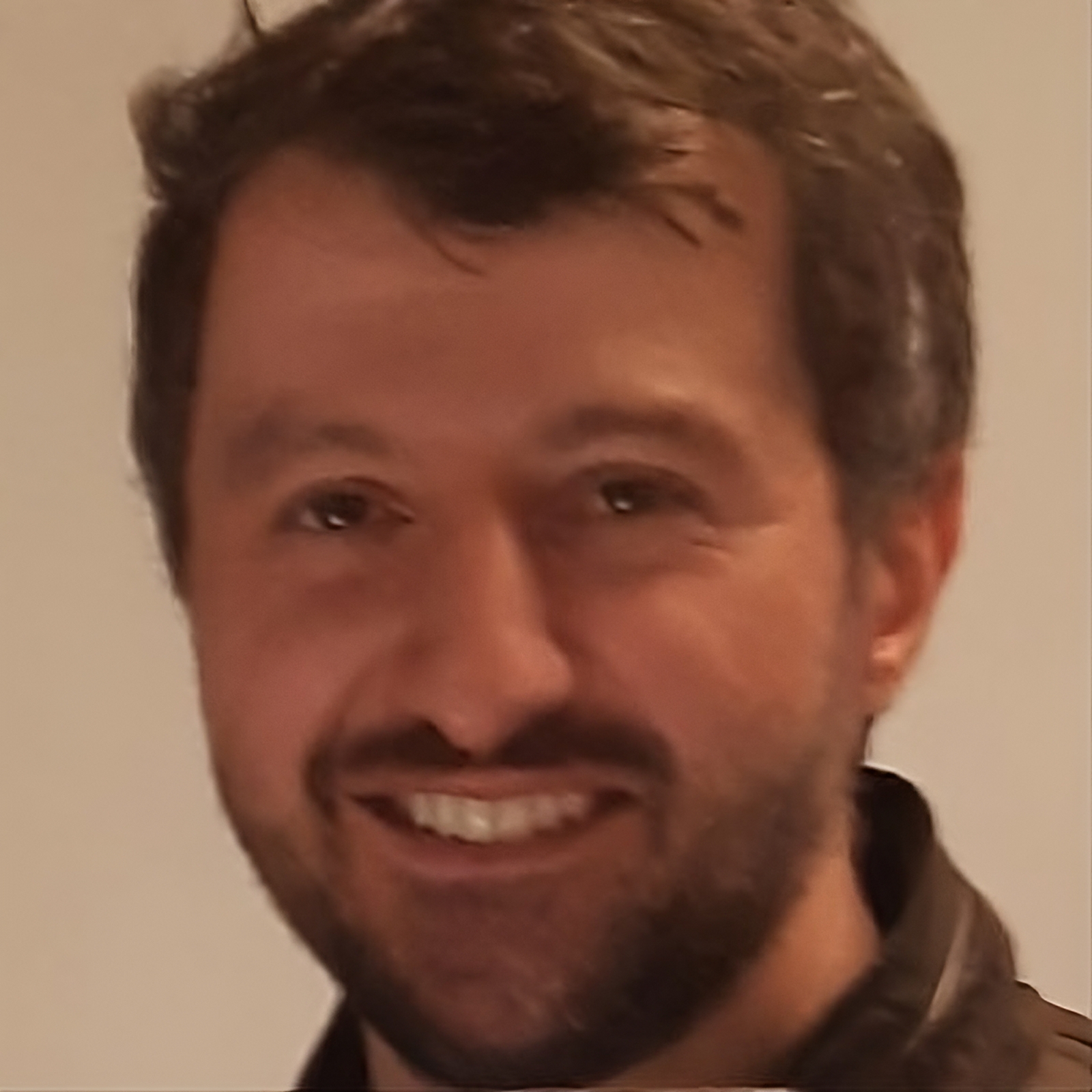}}]{Muhammet Balcilar}
joined InterDigital Inc. in 2021 where he is currently working as a Senior Scientist. He received
his B.S. and M.S. and PhD in computer engineering from Yildiz Technical University, Istanbul, Turkey in 2004, 2007 and 2013 respectively.  Previously, he worked as a postdoc at the university he graduated from, Ecole des Mines, Paris and University of 
Rouen Normandy in France. Both traditional and machine learning approaches in video compressing are his main field of research. He actively joins and proposes solutions to the JVET meetings. 
\end{IEEEbiography}
\vspace{-35pt}
\begin{IEEEbiography}
[{\includegraphics[width=1in,height=1.25in,clip,keepaspectratio]{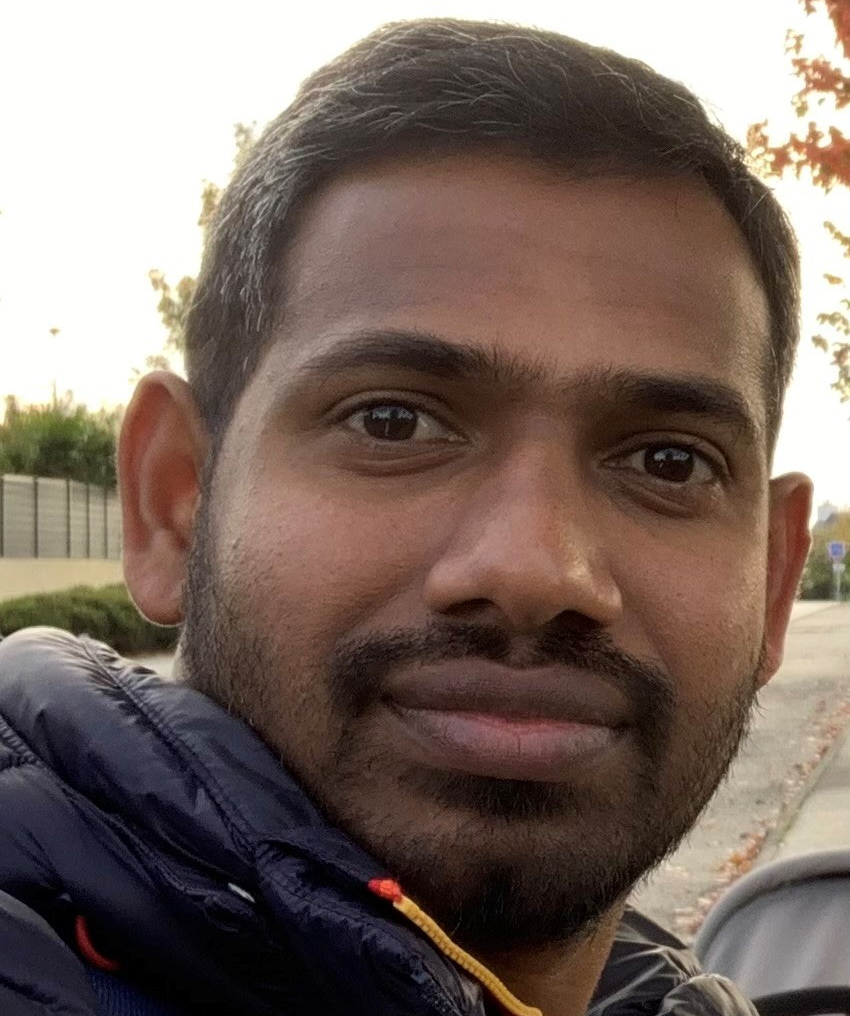}}]{Bharath Bhushan Damodaran}
 received the PhD degree from the Indian Institute of Space Science and Technology, Thiruvananthapuram, India, in 2015. He is currently a senior staff scientist with InterDigital Inc., France. In 2015-2020, he was a post-doctoral researcher with IRISA, Universite Bretagne Sud. He was awarded the Prestige and Marie Curie post-doctoral fellowship. His  current research interests includes machine learning approaches in image and video compression. 
\end{IEEEbiography}
\vspace{-35pt}
\begin{IEEEbiographynophoto}
{Karam Naser}
received his bachelor’s degree in electrical engineering from the University of Baghdad in Iraq in 2009. He obtained a master’s degree in communication engineering from RWTH Aachen University in Germany in 2013. He also holds a PhD degree from the University of Nantes in France in 2017. His PhD studies were conducted within the European Marie Curie Initial Training Network with the PROVISION project. He is currently working as a principal scientist in video compression at InterDigital Inc. He is actively involved in the standardization process and he is chairing multiple Ad-hoc groups and exploration experiment within the joint video exploration team (JVET).  
\end{IEEEbiographynophoto}
\vspace{-35pt}
\begin{IEEEbiography}
[{\includegraphics[width=1in,height=1.25in,clip,keepaspectratio]{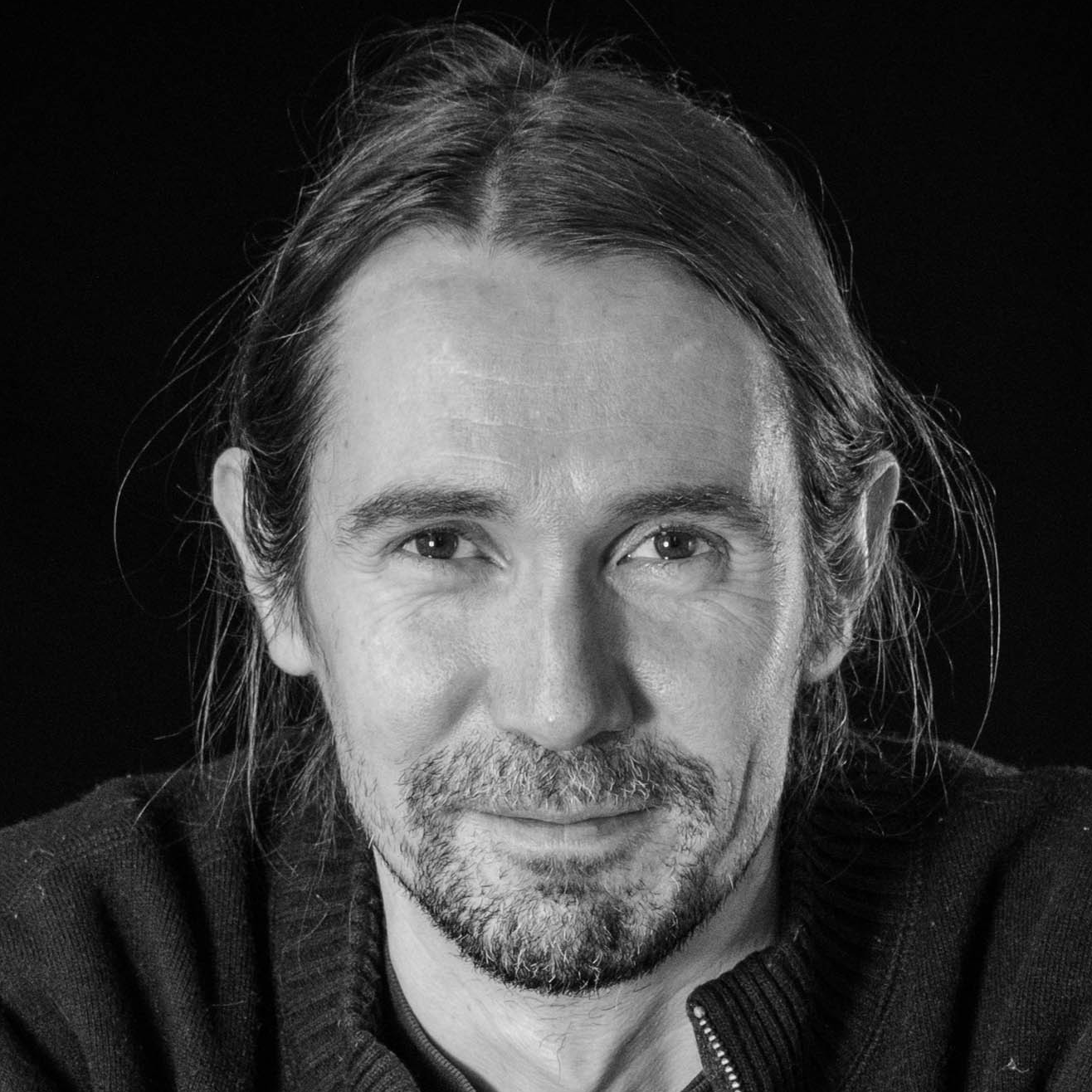}}]{Franck Galpin} received his M.Sc. in Engineering from IFSIC Rennes University in 1998, and a PhD degree in Computer Sciences and Signal Processing in 2002, from the same University. He has then spent 2 years as a JSPS post-doc at Tohoku University (Japan). In 2004, he joined the Toyota Higashi Fuji Research Center (Japan) where he has worked as a researcher on autonomous vehicle and robotics. He joined Technicolor R\&I in 2014, then InterDigital R\&I in 2019  where he is now working on video compression and deep learning for coding as a Senior Principal Scientist. He is an active participant of video codec standardization (VVC/H.266, next generation coding, Neural Network based
Video Coding), 
\end{IEEEbiography}
\vspace{-35pt}
\begin{IEEEbiography}[{\includegraphics[width=1in,height=1.25in,clip,keepaspectratio]{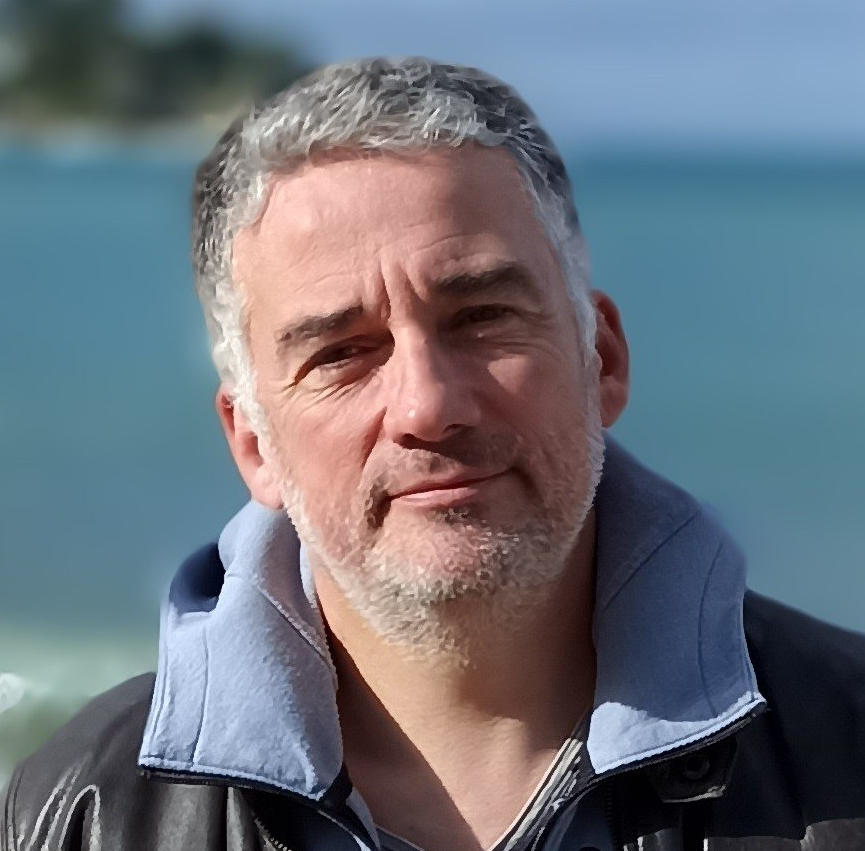}}]{Pierre Hellier} holds a engineering degree in aeronautics (Ensae), a MsC in applied mathematics (university of Toulouse, France), a PhD degree obtained in 2000 at Rennes University (France), and an habilitation thesis from Rennes university in 2010. After a postdoc at Utrecht university, Netherlands, he was appointed INRIA researcher in 2001, focusing on computer vision methods for medical image registration, image-guided neurosurgery and transcranial magnetic stimulation. In 2011, he moved to Technicolor research, working on UGC content synchronization, enhancement, and professional post-production. he pursued with InterDigital research in 2019, focusing on machine learning for image and video compression, as well as deep learning for digital human modeling. Since 2024, he is with the University of Rennes as associate professor. 
\end{IEEEbiography}



\end{document}